\documentclass[10pt,letterpaper,twocolumn]{article}
\usepackage[margin=0.75in]{geometry}
\usepackage{graphicx}
\usepackage{amsmath,amssymb,mathtools}
\usepackage[table]{xcolor}
\usepackage{booktabs,multirow}
\usepackage[numbers,sort&compress]{natbib}
\usepackage{caption}
\usepackage{algorithm}
\usepackage{algorithmic}
\usepackage{newfloat}
\usepackage{listings}
\usepackage{bibunits}
\usepackage[hyphens]{url}
\newcommand{\bestscore}[1]{\textbf{#1}}
\newcommand{\secondscore}[1]{\underline{#1}}
\newcommand{\thirdscore}[1]{\textit{#1}}
\definecolor{best}{HTML}{D32F2F}
\definecolor{second}{HTML}{1976D2}
\definecolor{third}{HTML}{388E3C}

\DeclareCaptionStyle{ruled}{labelfont=normalfont,labelsep=colon,strut=off}
\floatstyle{ruled}
\newfloat{listing}{tb}{lst}
\floatname{listing}{Listing}

\title{State-Isolated Vision Mamba via Update-Tendency Routing\\
for Underwater Image Enhancement}
\author{Yan Luo \and Laibin Chang \and Zefan Feng \and Yi'ang Chen
\and Kui Jiang \and Zhengguo Li}
\date{}
\begin{document}

\maketitle

\begin{abstract}
Vision Mamba has recently emerged as an efficient alternative to convolutional and Transformer-based architectures by modeling long-range spatial dependencies through recurrent state evolution. Existing studies primarily improve token serialization, scanning strategies, or expert routing, yet largely overlook a more fundamental question: how should recurrent states be organized for spatially heterogeneous visual restoration? In this paper, we show that forcing tokens with heterogeneous update behaviors to recurrently modify a shared hidden state leads to persistent cross-content interference, which we term State Contention. We further argue that effective recurrent modeling should organize state evolution according to token compatibility rather than token ordering. Based on this observation, we introduce State Isolation, a new recurrent state organization principle for Vision Mamba, where tokens with compatible state-update tendencies evolve within independent recurrent state streams. To instantiate this principle, we propose \textbf{IsoMamba}, a $\Delta$-guided architecture that constructs model-intrinsic update representations, discovers update-compatible token groups through label-free clustering, and routes them to isolated Mamba experts following a cluster--route--isolate paradigm. In addition, we develop a dynamic channel-wise reweighting objective to compensate for wavelength-dependent reconstruction difficulty in underwater images. Extensive experiments demonstrate that IsoMamba consistently achieves state-of-the-art performance on multiple underwater image enhancement benchmarks while providing substantially cleaner hidden-state responses, validating the effectiveness of state-isolated recurrent modeling. Our code is available at: https://github.com/kkoucy/IsoMamba.

\end{abstract}

\section{Introduction}
Underwater images suffer from wavelength-dependent absorption and scattering that accumulate along the line of sight~\cite{uie-formulation}.
Variations in scene depth, illumination, and water conditions make the resulting degradation spatially non-uniform~\cite{sea-thru}. Consequently, severely attenuated regions may require stronger color and contrast correction, whereas mildly degraded structures should be processed more conservatively. 
Effective underwater image enhancement (UIE) therefore requires spatially adaptive processing of heterogeneous restoration requirements.

CNN- and Transformer-based methods have significantly advanced UIE, yet modeling long range dependencies in high resolution images often involves a trade-off between computational cost and receptive field~\cite{UIEC-cnn,selTrans}. State Space Models (SSMs), particularly Mamba~\cite{mamba}, provide an efficient alternative by propagating information over long token sequences with linear complexity. Vision Mamba methods typically focus on how a two-dimensional feature grid is serialized, using raster, local, semantic, or frequency-aware scanning to improve information propagation~\cite{MambaIRv2,Fouriermamba,LocalMamba}. 
Recent restoration approaches further employ multi-expert routing to improve adaptability to diverse degradations~\cite{uni-uir,m2restore}. 
However, the organization of recurrent state sharing is typically a consequence of sequence construction or expert assignment rather than an explicit modeling objective.
Whether tokens that successively modify the same recurrent trajectory exhibit compatible state-update behavior remains largely unexamined.
This motivates a state-level question: \emph{which tokens should share the same recurrent state evolution?}


\begin{figure}[!t]
    \centering
    \includegraphics[width=0.98\linewidth]{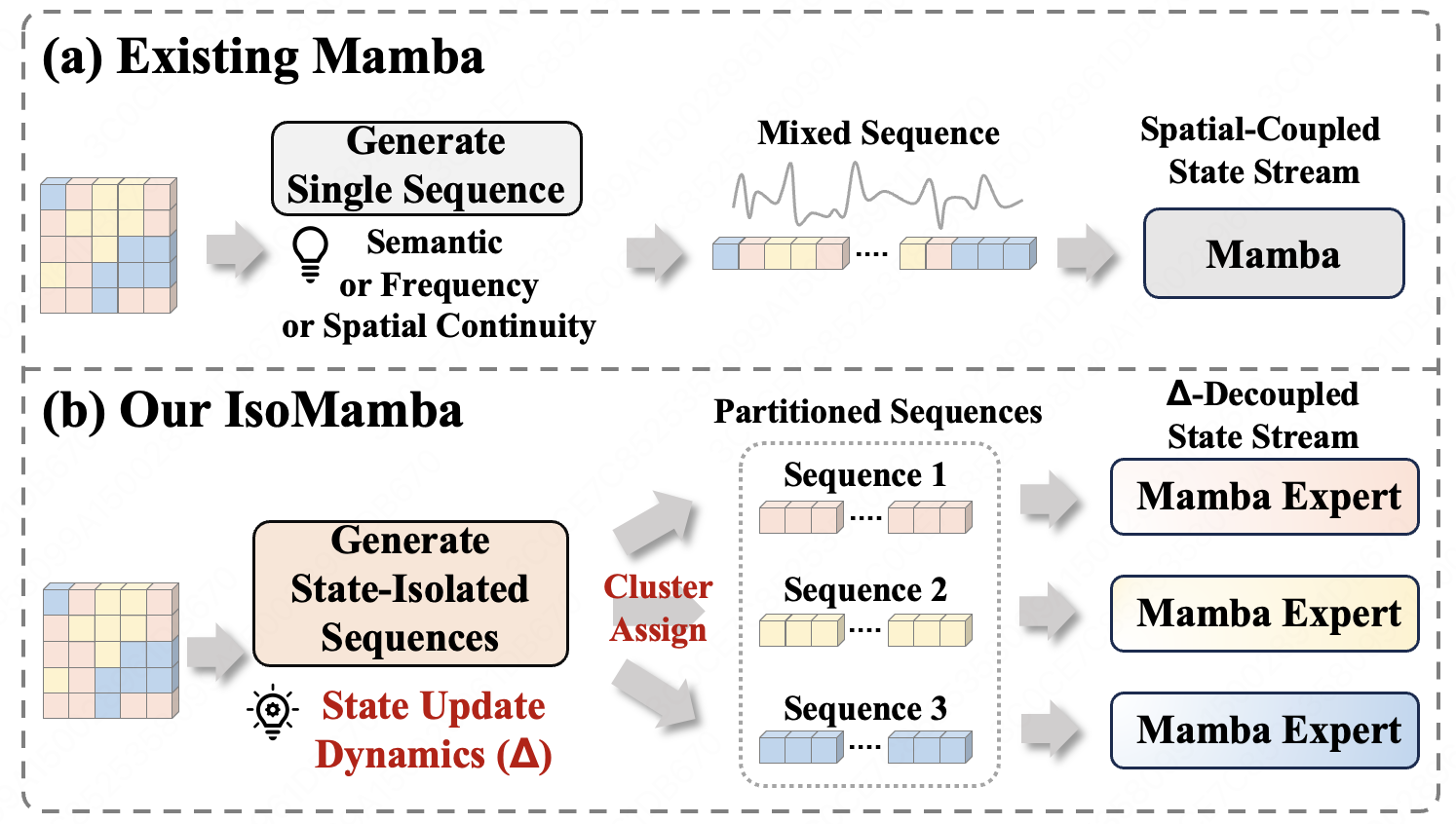}
    \caption{(a) Existing Mamba serializes features into a single mixed sequence with the spatial-coupled state stream. (b) IsoMamba clusters tokens by $\Delta$ and routes state-isolated sequences into decoupled Mamba experts.}
    \label{fig:m1}
    \vspace{-5mm}
\end{figure}

\begin{figure}[!tp]
    \centering
    \includegraphics[width=0.98\linewidth]{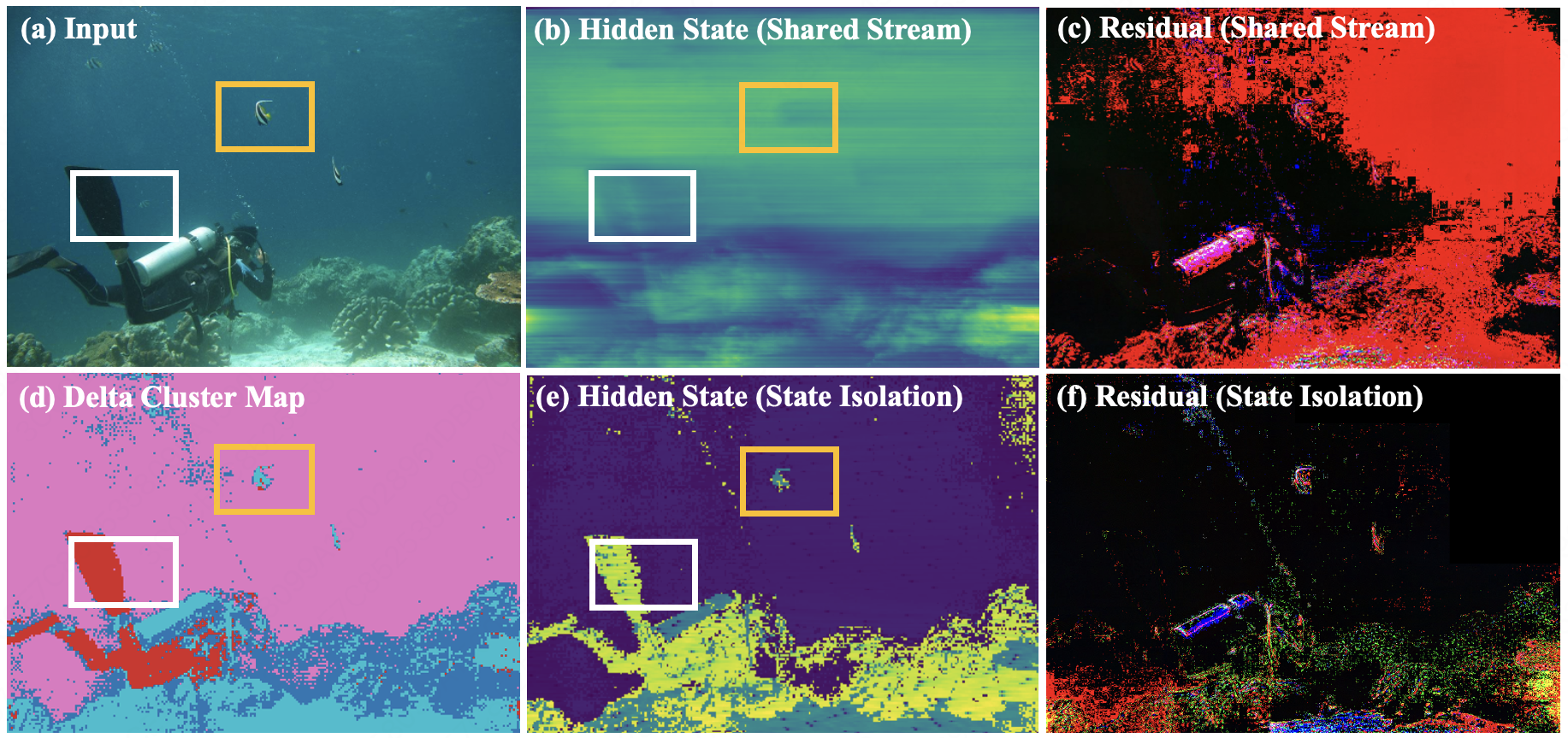}\vspace{-2mm}
\caption{
\textbf{State Contention versus State Isolation.}
\textbf{(a)} Input image.
\textbf{(d)} Token groups obtained from the $\Delta$-derived update representation.
\textbf{(b), (e)} Hidden-state responses of the shared-state baseline and IsoMamba, respectively.
\textbf{(c), (f)} Their corresponding channel-wise absolute reconstruction residuals using same normalization range.
}
    \label{fig:m2}\vspace{-4mm}
\end{figure}

We investigate this question in the context of spatially heterogeneous image restoration. 
After serialization, tokens associated with different image contents successively update a shared recurrent state stream, $h_t=\bar{A}_t h_{t-1}+\bar{B}_t x_t$.
When tokens with dissimilar state-update tendencies successively modify the same recurrent state, their responses are accumulated within a common trajectory, which may obscure group-specific state information.
We refer to this forced recurrent mixing as \textbf{State Contention}.
Unlike generic feature entanglement, State Contention specifically refers to the coupling introduced during recurrent evolution, where heterogeneous tokens sequentially modify a shared state. 
Figure~\ref{fig:m2}(b)--(c) show that the shared-state baseline produces weakly structured hidden-state responses and widespread reconstruction residuals.
Although token reordering changes local dependencies, it does not alter the underlying state-sharing relationship.

To reorganize state sharing, we require a model-intrinsic criterion for estimating token compatibility. 
In Mamba, the input-dependent discretization signal $\Delta$ modulates the discretized state transition and input injection~\cite{mamba}. 
We therefore construct a $\Delta$-derived representation as a proxy for each token's \textbf{state-update tendency} and group tokens by similarity in this space.
The coherent patterns in Figure~\ref{fig:m2}(d) suggest that these groups capture structured update behavior without relying on explicit semantic or spatial criteria. 
Based on this grouping, we introduce \textbf{State Isolation}: tokens with similar estimated update tendencies share recurrent evolution, whereas different groups evolve through separate state streams.
This design is intended to resolve interference between tokens with dissimilar update behaviors while preserving recurrent interaction within each compatible group.

We instantiate this principle with \textbf{IsoMamba}, a $\Delta$-guided architecture following a \emph{cluster--route--isolate} paradigm.
IsoMamba groups tokens in the learned $\Delta$-derived space and routes each group to a dedicated Mamba expert with an independent recurrent state. The resulting features are remapped to their original spatial layout and fused with lightweight global context to preserve image-level coherence. 
As shown in Figure~\ref{fig:m2}(e)--(f), IsoMamba produces more structured hidden-state responses and lower reconstruction residuals than the shared-state baseline, consistent with reduced State Contention. 
Furthermore, to address wavelength-dependent differences in restoration difficulty, we introduce a dynamic channel-wise reweighting objective that adjusts RGB supervision according to the current reconstruction errors.

Our main contributions are summarized as follows:
\begin{itemize}
    \item We identify \textbf{State Contention}, a previously overlooked limitation of Vision Mamba in which tokens with heterogeneous update requirements repeatedly modify a shared recurrent state, causing cross-content interference during spatially heterogeneous restoration.

    \item We introduce \textbf{State Isolation} and instantiate it with \textbf{IsoMamba}, a $\Delta$-guided architecture that constructs model-intrinsic state-update representations and follows a label-free \emph{cluster--route--isolate} paradigm, routing tokens with similar estimated update tendencies without semantic supervision or hand-crafted partitions.

    \item We develop a dynamic channel-wise reweighting objective that adaptively balances wavelength-dependent reconstruction difficulty, complementing State Isolation to alleviate residual color bias in UIE task.
\end{itemize}

\begin{figure*}[t]
    \centering
    \includegraphics[width=\textwidth]{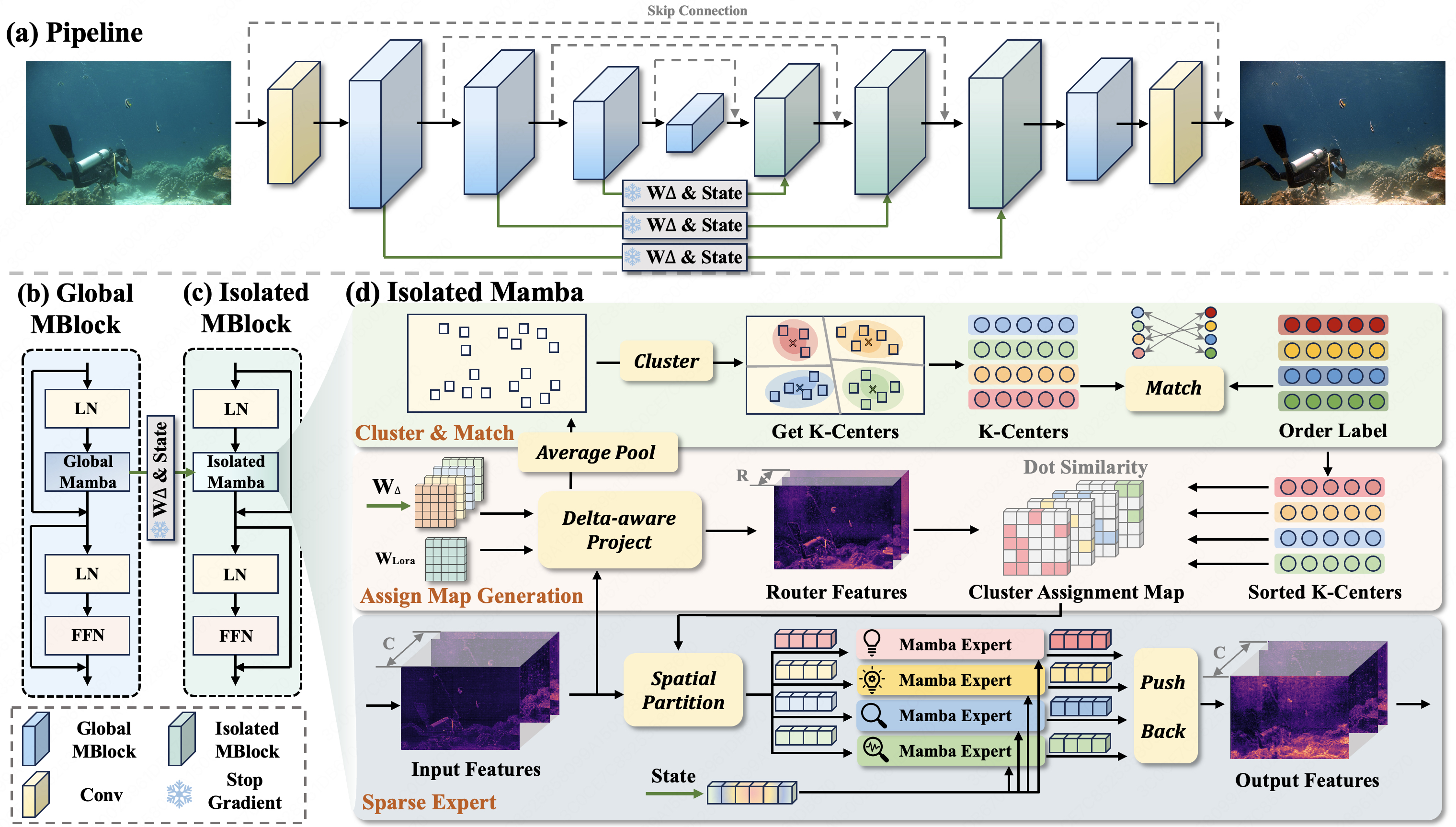}
\caption{
\textbf{Overview of IsoMamba.}
The proposed framework consists of two complementary stages.
\textbf{(a)} Overall architecture.
\textbf{(b)} Global recurrent modeling establishes image-level context through a shared state stream.
\textbf{(c)} State-Isolation MBlock reorganizes recurrent state evolution by separating tokens with incompatible update tendencies into independent state streams.
\textbf{(d)} A cluster--route--isolate strategy realizes State Isolation using model-intrinsic update tendency derived from $\Delta$.
}
    \label{pipeline}
    \vspace{-4mm}
\end{figure*}


\section{Methodology}
\subsection{Overview}

As illustrated in Figure~\ref{pipeline}, IsoMamba adopts an asymmetric U-shaped architecture. The encoder uses Global MBlocks with four direction selective scanning to establish image level context, while the decoder employs Isolated MBlocks to reorganize recurrent evolution according to token-wise update compatibility. At each stage, the encoder provides the corresponding decoder stage with two stop-gradient cues: the input-dependent discretization signal $\Delta$ for update-aware routing and the aggregated terminal states for global coordination. Each decoder stage then performs update-tendency modeling, state-compatible routing, and isolated recurrent evolution.

\subsection{State-Update Representation}

State Isolation requires estimating which tokens can compatibly share recurrent evolution. For a selective state-space model, the token-wise update is
\[
h_i=\exp(\Delta_i A)\odot h_{i-1}
    +{B}_i(\Delta_i\odot x_i).
\tag{1}
\]
Because the eigenvalues of $A$ are negative, $\Delta_i$ modulates the trade-off between historical-state retention and input-driven renewal. We therefore use its multidimensional response as a model-intrinsic proxy for the token's state-update tendency.

Given a decoder feature map $X\in\mathbb{R}^{B\times H\times W\times C}$, we unfold it along four scan directions and reuse the corresponding encoder update projections:
\[
R_{\Delta}
=
\operatorname{Concat}_{d=1}^{4}
[
\mathcal{T}_{d}^{-1}
\bigl(X^{(d)}sg[W_{\Delta}^{(d)}]\bigr)],
\tag{2}
\]
where $\mathcal{T}_{d}^{-1}$ restores the spatial layout and $\operatorname{sg}[\cdot]$ denotes stop-gradient. To account for the encoder decoder distribution gap, we introduce a low-rank residual adaptation:
\[
R=\operatorname{Softplus}
\left(R_{\Delta}+\alpha(XW_A)W_B\right),
\tag{3}
\]
where $W_A$ and $W_B$ are low-rank matrices and $\alpha$ is learnable. The resulting representation $R\in\mathbb{R}^{B\times H\times W\times D}$ is used to estimate similarity in recurrent update behavior rather than explicit semantic or spatial similarity.

\subsection{Cluster--Route--Isolate}

Given the update-tendency representation $R$, IsoMamba constructs an instance-adaptive partition of $K$ token groups. To reduce routing overhead, clustering is performed on an $\ell_2$-normalized and spatially pooled representation, and the centers are updated from their assigned tokens within each forward pass. The resulting partition therefore adapts to the input-specific update structure.

The instance-specific centers are unordered: although the set
$\mathcal{C}=\{c_m\}_{m=1}^{K}$ is meaningful, its indices may permute across images and training iterations. Directly associating these indices with fixed experts would expose an expert to inconsistent update regimes. We therefore introduce a fixed set of orthonormal anchors
$\mathcal{P}=\{p_k\}_{k=1}^{K}$, initialized once and kept unchanged throughout training, where $p_i^\top p_j=\delta_{ij}$. For each image, we compute
\[
S_{mk}=\widehat{c}_m^\top p_k
\tag{4}
\]
and perform greedy one-to-one maximum similarity matching. Let $\pi(k)$ denote the center matched to anchor $p_k$; the ordered center associated with expert $k$ is then
\[
\widetilde{c}_k = l_2norm(c_{\pi(k)}).
\tag{5}
\label{eq:center_matching}
\]
This anchor referenced ordering provides persistent order labels for the experts and encourages consistent expert assignments across instances. 

Using the ordered centers, we compute full-resolution routing probabilities as
\[
A_{hwk}
=
\operatorname{Softmax}_{k}
\left(
\frac{
{R}_{hw}{\widetilde{c}}_k^{\top}
}{\tau}
\right),
\tag{6}
\label{eq:soft_routing}
\]
where $\tau=\exp(\theta)$ is a learnable positive temperature. A Top-1 assignment is used in the forward pass, while a straight-through estimator propagates gradients through the soft probabilities:
\[
M^{\mathrm{hard}}_{hw}
=
\operatorname{OneHot}
\left(
\arg\max_k A_{hwk}
\right),
\]
\[
M
=
\operatorname{sg}
\left[
M^{\mathrm{hard}}-A
\right]+A,
\tag{7}
\label{eq:ste_routing}
\]
Thus, $M$ equals the binary Top-1 assignment in the forward pass, while its gradients follow $A$.

Unlike conventional expert routing, the proposed assignment determines not only which expert processes a token, but also which tokens enter the same recurrent state stream. Clustering estimates state-update compatibility, while anchor-based ordering encourages a consistent association between update regimes and expert identities across instances.

\subsection{Isolated Recurrent Evolution}

Tokens assigned to group $k$ are gathered into a compact sequence and processed by a dedicated Mamba expert:
\[
Y_k
=
\operatorname{Mamba}_k
\left(
X_{hw}\mid M_{hw}=1
\right).
\tag{8}
\]
Each expert maintains an independent recurrent state, allowing tokens with similar estimated update tendencies to evolve within the same stream while isolating different groups. This organization is designed to alleviate State Contention by preventing dissimilar update regimes from repeatedly modifying a common recurrent trajectory. The expert outputs are subsequently pushed back to their original coordinates and merged into the decoder feature map, preserving spatial correspondence and output resolution.

\subsection{Global State Coordination}
State isolation improves local adaptivity but may weaken communication across groups. We therefore inject a shared global state into each expert through lightweight feature-wise modulation.

At each decoder stage, the terminal hidden states from the four encoder scan directions are aggregated into a global descriptor $H_{\mathrm{g}}$. For an expert output $Y_k$, channel-wise modulation parameters are predicted as
\begin{equation}
[\gamma,\beta]
=
\operatorname{MLP}(sg[H_{\mathrm{g}}]),
\qquad
\widetilde{Y}_k
=
Y_k\odot(1+\gamma)+\beta.
\label{eq:merged-1}
\tag{9}
\end{equation}

The same global descriptor is shared across experts, coordinating independently restored regions under a common image-level context. The final MLP layer is zero-initialized so that the modulation begins as an identity mapping.

\begin{figure}[!tbp]
    \centering
    \includegraphics[width=1\linewidth]{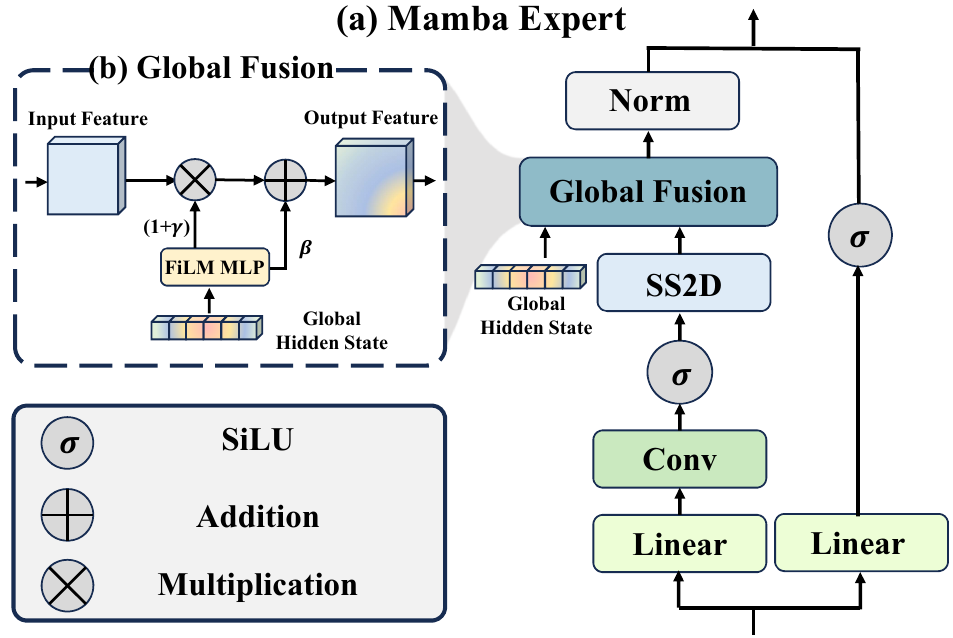}\vspace{-2mm}
    \caption{
    The detailed structure of Mamba expert.
    }
    \label{fig:global_fusion}\vspace{-3mm}
\end{figure}

\subsection{Objective Function}
State Isolation addresses spatially heterogeneous restoration, while underwater degradation is also strongly wavelength-dependent. Since RGB channels may exhibit markedly different reconstruction difficulty, a channel-symmetric objective can be dominated by easier channels.

For each channel $c\in\{R,G,B\}$, we compute its current reconstruction error:
\begin{equation}
E_c
=
\frac{1}{HW}
\sum_x
\left|
\hat{I}_c(x)-I_c(x)
\right|.
\tag{10}
\end{equation}

The corresponding difficulty weight is
\begin{equation}
w_c
=
\alpha
+
(1-\alpha)
\frac{E_c}{\bar{E}+\epsilon},
\qquad
\bar{E}
=
\frac{1}{3}\sum_c E_c,
\tag{11}
\label{eq:merged-2}
\end{equation}
where gradients through $w_c$ are stopped. The channel-adaptive reconstruction loss is then
\begin{equation}
\mathcal{L}_{\mathrm{ca}}
=
\frac{1}{HW}
\sum_c\sum_x
w_c
\left|
\hat{I}_c(x)-I_c(x)
\right|.
\tag{12}
\end{equation}

The complete training objective combines channel-adaptive reconstruction, structural similarity, and edge preservation:
\begin{equation}
\mathcal{L}
=
\lambda_1\mathcal{L}_{\mathrm{ca}}
+
\lambda_2\mathcal{L}_{\mathrm{ssim}}
+
\lambda_3\mathcal{L}_{\mathrm{edge}},
\tag{13}
\end{equation}
where $\lambda_1=8$, $\lambda_2=1$, and $\lambda_3=4$.

\begin{table*}[!t]
\centering
\caption{
\textbf{Quantitative comparison on the full-reference datasets UIEB, LSUI, and EUVP.}
The best, second-best, and third-best results are shown in
\textbf{bold}, \underline{underlined}, and \textit{italic} type, respectively.
}
\label{tab:full-ref}
\vspace{-2mm}
\renewcommand{\arraystretch}{0.95}
\setlength{\tabcolsep}{3.5pt}
\resizebox{\textwidth}{!}{
\begin{tabular}{@{}ll ccc ccc ccc ccc@{}}
\toprule
\multirow{2}{*}{Method} &
\multirow{2}{*}{Venue} &
\multicolumn{3}{c}{\textbf{UIEB}} &
\multicolumn{3}{c}{\textbf{LSUI}} &
\multicolumn{3}{c}{\textbf{EUVP}} &
\multicolumn{3}{c}{\textbf{Complexity}} \\
\cmidrule(lr){3-5}
\cmidrule(lr){6-8}
\cmidrule(lr){9-11}
\cmidrule(lr){12-14}
& & PSNR$\uparrow$ & SSIM$\uparrow$ & MSE$\downarrow$
& PSNR$\uparrow$ & SSIM$\uparrow$ & MSE$\downarrow$
& PSNR$\uparrow$ & SSIM$\uparrow$ & MSE$\downarrow$
& Params (M)$\downarrow$ & FLOPs (G)$\downarrow$ & Time (s)$\downarrow$ \\
\midrule
U-shape
& TIP23
& 20.54 & 0.772 & 91.77
& 24.16 & 0.846 & 81.40
& 25.48 & 0.830 & 74.20
& 65.60 & 66.20 & 0.063 \\

DM-UIE
& MM23
& 23.46 & 0.861 & 76.30
& 27.27 & 0.849 & 66.66
& 26.86 & 0.865 & 57.04
& 10.70 & 66.89 & 0.197 \\

GUPDM
& MM23
& 23.99 & 0.877 & 75.49
& 27.67 & 0.867 & 56.97
& 26.21 & 0.850 & 64.45
& \bestscore{1.49} & 95.80 & 0.696 \\

Semi-UIR
& CVPR23
& 23.49 & 0.860 & 79.86
& 27.46 & 0.878 & 63.57
& 26.85 & 0.876 & 60.57
& \secondscore{3.31} & 72.88 & 0.785 \\

wMamba
& arXiv24
& 23.55 & 0.888 & 78.32
& 28.71 & 0.894 & 50.32
& 28.55 & 0.882 & 47.01
& \thirdscore{3.69} & \bestscore{7.53} & \secondscore{0.028} \\

CECF
& AAAI24
& 21.94 & 0.868 & 86.06
& 26.34 & 0.878 & 64.94
& 28.34 & 0.874 & 49.35
& 39.69 & 83.58 & \bestscore{0.002} \\

MambaIR
& ECCV24
& 24.04 & 0.912 & 75.99
& 29.16 & 0.903 & 47.97
& 29.23 & 0.892 & 43.07
& 25.92 & 137.57 & 0.103 \\

WF-Diff
& CVPR24
& 22.78 & 0.912 & 83.64
& 27.49 & 0.894 & 59.79
& 27.01 & 0.876 & 60.06
& 18.48 & 570.09 & 0.146 \\

HCLR
& IJCV24
& 21.98 & 0.887 & 84.74
& 28.74 & 0.889 & 49.93
& 27.86 & 0.864 & 52.38
& 4.87 & 401.90 & \thirdscore{0.054} \\

F-RWKV
& arXiv25
& 24.30 & 0.912 & 76.58
& 28.81 & 0.899 & 50.90
& 29.43 & 0.882 & 42.85
& 5.31 & \thirdscore{15.69} & 0.088 \\

HINT
& ICCV25
& 24.64 & \secondscore{0.922} & 73.88
& 29.31 & \secondscore{0.910} & 47.87
& 29.14 & 0.897 & 45.36
& 24.73 & 142.12 & 0.080 \\

Concertormer
& ICCV25
& 22.55 & 0.901 & 84.27
& 26.97 & 0.893 & 60.60
& 28.53 & \secondscore{0.901} & 48.07
& 45.96 & 201.01 & 0.108 \\

MaIR
& CVPR25
& \secondscore{25.11} & \bestscore{0.923} & \thirdscore{71.33}
& \secondscore{29.76} & \thirdscore{0.908} & \secondscore{44.70}
& \secondscore{29.90} & \thirdscore{0.899} & \secondscore{39.80}
& 7.78 & \secondscore{12.59} & 0.377 \\

EVSSM
& CVPR25
& \thirdscore{24.93} & 0.916 & \secondscore{70.09}
& \thirdscore{29.49} & 0.902 & \thirdscore{45.97}
& \thirdscore{29.67} & 0.885 & \thirdscore{41.31}
& 17.10 & 126.00 & 0.279 \\

DTIUIE
& TIP26
& 24.57 & 0.911 & 71.39
& 28.18 & 0.893 & 53.89
& 28.10 & 0.880 & 49.64
& 50.81 & 385.93 & 0.112 \\
\midrule
\rowcolor{black!8}
\textbf{IsoMamba (Ours)}
& --
& \bestscore{25.32} & \thirdscore{0.920} & \bestscore{68.21}
& \bestscore{29.89} & \bestscore{0.914} & \bestscore{43.64}
& \bestscore{30.09} & \bestscore{0.902} & \bestscore{33.41}
& 15.23 & 51.59 & 0.095 \\
\bottomrule
\end{tabular}
}
\vspace{-3mm}
\end{table*}

\begin{table*}[!t]
\centering
\caption{
\textbf{Quantitative comparison on the no-reference datasets U45, C60, and SQUID.}
The best, second-best, and third-best results are shown in
\textbf{bold}, \underline{underlined}, and \textit{italic} type, respectively.
Tied values are assigned the same rank.
}
\label{tab:no-ref}
\vspace{-2mm}
\renewcommand{\arraystretch}{0.95}
\setlength{\tabcolsep}{3.5pt}
\resizebox{\textwidth}{!}{
\begin{tabular}{@{}ll cccc cccc cccc@{}}
\toprule
\multirow{2}{*}{Method} &
\multirow{2}{*}{Venue} &
\multicolumn{4}{c}{\textbf{U45}} &
\multicolumn{4}{c}{\textbf{C60}} &
\multicolumn{4}{c}{\textbf{SQUID}} \\
\cmidrule(lr){3-6}
\cmidrule(lr){7-10}
\cmidrule(lr){11-14}
& & UCIQE$\uparrow$ & UIQM$\uparrow$ & URanker$\uparrow$ & CLIP-IQA$\uparrow$
& UCIQE$\uparrow$ & UIQM$\uparrow$ & URanker$\uparrow$ & CLIP-IQA$\uparrow$
& UCIQE$\uparrow$ & UIQM$\uparrow$ & URanker$\uparrow$ & CLIP-IQA$\uparrow$ \\
\midrule
U-shape
& TIP23
& 0.567 & 3.105 & 1.242 & 0.435
& 0.545 & 2.660 & 0.956 & 0.457
& 0.544 & 2.230 & 0.554 & 0.524 \\

DM-UIE
& MM23
& 0.603 & 3.108 & 1.810 & 0.537
& 0.580 & 2.577 & 1.341 & 0.559
& \bestscore{0.659} & 0.519 & -0.070 & 0.560 \\

GUPDM
& MM23
& 0.573 & 3.028 & 1.591 & 0.528
& 0.550 & 2.591 & 1.150 & 0.556
& 0.577 & 2.486 & 0.952 & 0.568 \\

Semi-UIR
& CVPR23
& 0.605 & 3.131 & \secondscore{2.031} & 0.510
& \thirdscore{0.585} & 2.648 & \bestscore{1.596} & 0.556
& \secondscore{0.588} & 2.309 & \secondscore{1.134} & 0.586 \\

wMamba
& arXiv24
& 0.608 & 3.068 & 2.026 & 0.531
& 0.580 & \secondscore{2.773} & 1.415 & 0.504
& 0.576 & \secondscore{2.673} & 1.003 & 0.558 \\

CECF
& AAAI24
& 0.606 & 3.096 & 1.965 & 0.505
& 0.584 & 2.695 & 1.440 & 0.515
& 0.569 & 2.369 & 0.881 & 0.480 \\

MambaIR
& ECCV24
& 0.613 & 3.091 & 2.006 & 0.547
& 0.582 & 2.679 & 1.442 & \bestscore{0.583}
& 0.565 & 2.482 & 0.895 & \bestscore{0.677} \\

WF-Diff
& CVPR24
& 0.612 & 3.068 & \thirdscore{2.029} & 0.537
& 0.584 & 2.715 & 1.463 & 0.538
& 0.580 & 2.572 & 0.806 & 0.651 \\

HCLR
& IJCV24
& 0.592 & \bestscore{3.168} & 1.729 & \thirdscore{0.554}
& 0.566 & \thirdscore{2.737} & 1.333 & \thirdscore{0.573}
& 0.576 & 2.449 & 0.947 & 0.577 \\

F-RWKV
& arXiv25
& \bestscore{0.616} & 3.055 & 1.969 & 0.544
& \bestscore{0.589} & 2.510 & 1.433 & \secondscore{0.574}
& 0.569 & 2.403 & 0.689 & 0.614 \\

HINT
& ICCV25
& 0.613 & \thirdscore{3.136} & \bestscore{2.069} & 0.542
& 0.578 & 2.614 & \thirdscore{1.493} & 0.569
& 0.578 & 2.529 & 0.937 & \secondscore{0.671} \\

Concertormer
& ICCV25
& 0.597 & 3.088 & 1.792 & \secondscore{0.556}
& 0.563 & 2.632 & 1.219 & \secondscore{0.574}
& 0.546 & 2.390 & 0.556 & \thirdscore{0.663} \\

MaIR
& CVPR25
& 0.612 & 3.083 & 1.875 & 0.541
& \bestscore{0.589} & 2.564 & 1.375 & 0.562
& \thirdscore{0.586} & 2.473 & 0.928 & 0.653 \\

EVSSM
& CVPR25
& \secondscore{0.615} & 3.100 & 1.971 & 0.542
& 0.584 & 2.640 & 1.422 & 0.564
& 0.551 & 2.114 & 0.576 & 0.611 \\

DTIUIE
& TIP26
& 0.572 & 2.650 & 1.333 & 0.533
& 0.496 & \bestscore{2.892} & 1.128 & 0.518
& 0.564 & \bestscore{3.071} & \bestscore{1.583} & 0.420 \\
\midrule
\rowcolor{black!8}
\textbf{IsoMamba (Ours)}
& --
& \thirdscore{0.614} & \secondscore{3.142} & 1.933 & \bestscore{0.596}
& \secondscore{0.586} & 2.621 & \secondscore{1.547} & 0.554
& 0.584 & \thirdscore{2.586} & \thirdscore{1.005} & 0.604 \\
\bottomrule
\end{tabular}
}
\vspace{-3mm}
\end{table*}

\section{Experiments}

\subsection{Experimental Settings}
\textbf{Datasets.}
We jointly train IsoMamba from scratch on the union of the UIEB~\cite{C60}  and LSUI~\cite{lsui} training sets, containing 800 and 3,502 paired images, respectively.
Evaluation is conducted under both full-reference (FR) and no-reference (NR) settings.
For FR evaluation, we use 90 images from UIEB, 400 images from LSUI, and 481 images from EUVP~\cite{EUVP}.
For NR evaluation, we use 45 images from U45~\cite{U45}, 60 images from C60~\cite{C60}, and 16 images from SQUID~\cite{squid}.
Following a unified resize-based protocol, all test images are resized to $256\times256$ for every competing method.

\textbf{Metrics.}
For paired benchmarks, we report PSNR, SSIM, and MSE to measure pixel fidelity and structural preservation.
For unpaired real-world benchmarks, we report UCIQE~\cite{uciqe}, UIQM~\cite{uiqm}, URanker~\cite{Uranker}, and CLIP-IQA~\cite{clip-iqa}, which assess underwater-specific visibility, color naturalness, perceptual quality, and semantic image quality from complementary perspectives.

\textbf{Implementation Details.}
IsoMamba is implemented in PyTorch 2.4.1 and trained on a single NVIDIA RTX 4090 GPU.
We use AdamW with an initial learning rate of $1\times10^{-4}$ and $(\beta_1,\beta_2)=(0.9,0.999)$.
The network is optimized for 300K iterations with a batch size of 2, using cosine annealing for learning-rate decay.
Unless otherwise stated, all ablations are conducted on the UIEB test set with identical training configurations.


\textbf{Compared Methods.} 
We compare IsoMamba with representative CNN-, Transformer-, diffusion-, RWKV-, and Mamba-based methods, including U-Shape~\cite{lsui}, DM-UIE~\cite{Tang2023UnderwaterIE}, GUPDM~\cite{GUPDM}, Semi-UIR~\cite{Semi-UIE}, wMamba~\cite{guan2024watermamba}, CECF~\cite{CECF}, MambaIR~\cite{MambaIR}, WF-Diff~\cite{WD-Diff}, HCLR~\cite{HCLR}, F-RWKV~\cite{Fourier-rwkv}, HINT~\cite{HINT}, Concertormer~\cite{Concertormer}, MaIR~\cite{MaIR}, EVSSM~\cite{EVSSM} and DTIUIE\cite{dtiuie}.

\begin{figure*}
    \centering
    \includegraphics[width=1\linewidth]{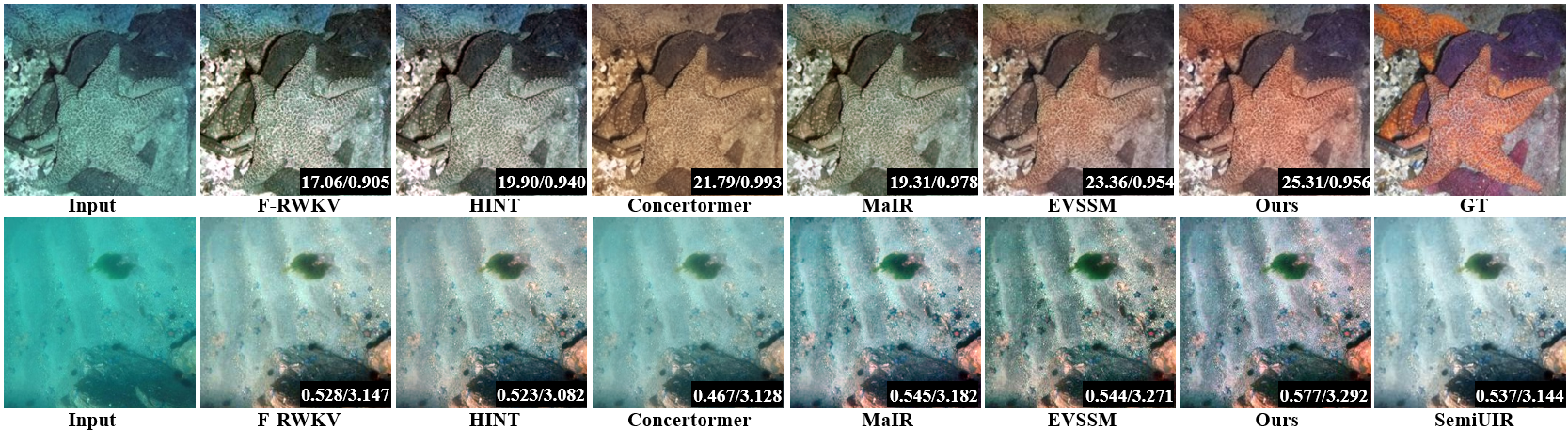}\vspace{-2mm}
    \caption{
    \textbf{Visual comparison under paired and unpaired settings.}
    In the paired example, IsoMamba recovers colors and fine structures that more closely agree with the reference.
    In the unpaired example, it suppresses severe color casts without introducing oversaturation or texture artifacts.
    }
    \label{fig:visual}\vspace{-4mm}
\end{figure*}

\subsection{Comparisons with State-of-the-Arts}


\textbf{Full-Reference Performance.}
Tables~\ref{tab:full-ref} shows that IsoMamba achieves the highest PSNR and lowest MSE on all three paired benchmarks.
It obtains 25.32,dB on UIEB, 29.89,dB on LSUI, and 30.09,dB on EUVP, demonstrating that the improvement is consistent across different data distributions rather than being tied to a single training domain.
The gains are particularly meaningful because IsoMamba does not rely on globally increasing model capacity.
Instead, it reorganizes recurrent state evolution so that tokens with incompatible update requirements no longer repeatedly overwrite the same state stream. IsoMamba also maintains a favorable computational profile, using 15.23M parameters, 51.59 GFLOPs, and 0.095s per $256\times256$ image.
This result indicates that State Isolation preserves the linear token-processing property of Mamba and improves the use of recurrent capacity rather than simply introducing a substantially larger restoration network.

\textbf{Real-World Perceptual Performance.}
Table~\ref{tab:no-ref} reports results on three unpaired real-world benchmarks.
Although underwater no-reference metrics are known to emphasize different and occasionally conflicting perceptual properties, IsoMamba remains competitive across all four metrics.
It achieves the best CLIP-IQA score on U45 and ranks among the top-performing methods on the remaining datasets and metrics.
The consistency between FR and NR results suggests that State Isolation improves both reference-based reconstruction fidelity and perceptual restoration under unknown real-world degradation.


\textbf{Visual Comparison.}
Figure~\ref{fig:visual}  further demonstrates the advantage of organizing recurrent computation according to update compatibility.
Methods based on globally shared restoration pathways often under-correct severely degraded regions or over-enhance already visible structures.
In contrast, IsoMamba performs differentiated restoration across spatial regions: heavily degraded areas receive stronger renewal, while reliable structures retain more historical information.
This behavior produces more natural colors, sharper local details, and fewer artificial boundaries.

\textbf{Full Resolution Evaluation.}
The main comparisons follow a unified 256 × 256 protocol to ensure consistent inputs across methods. We further evaluate representative models on full-resolution UIEB and LSUI images using the same RTX 4090 GPU. Models exceeding the 24 GB memory limit use 512 × 512 tiles with a 64-pixel overlap, while IsoMamba processes entire images without tiling. Figure 6 shows that full-image processing preserves global color consistency and avoids boundary artifacts associated with patch-wise inference, demonstrating the applicability of IsoMamba to megapixel underwater images.

\begin{table}[t]
\centering
\caption{
\textbf{Full resolution evaluation.}
Entries are PSNR$\uparrow$/SSIM$\uparrow$/UCIQE$\uparrow$/UIQM$\uparrow$.
$\dagger$ denotes tiled inference while others use full-image inference.
}
\vspace{-2mm}
\footnotesize
\setlength{\tabcolsep}{3pt}
\begin{tabular}{@{}l|cc@{}}
\toprule
Method & UIEB & LSUI \\
\midrule
HINT$^\dagger$
& 21.63/0.881/0.604/2.888
& 27.44/0.897/0.597/2.880 \\
EVSSM$^\dagger$
& 20.72/0.875/0.604/\textbf{2.935}
& 26.75/0.883/0.593/2.835 \\
MaIR
& 23.19/0.892/0.605/2.910
& 28.53/0.907/0.596/\textbf{2.911} \\
\textbf{Ours}
& \textbf{23.94/0.895/0.615}/2.849
& \textbf{29.01/0.909/0.599}/2.881 \\
\bottomrule
\end{tabular}
\label{tab:full_resolution}
\end{table}

\begin{figure}[h]
    \centering
    \includegraphics[width=1\linewidth]{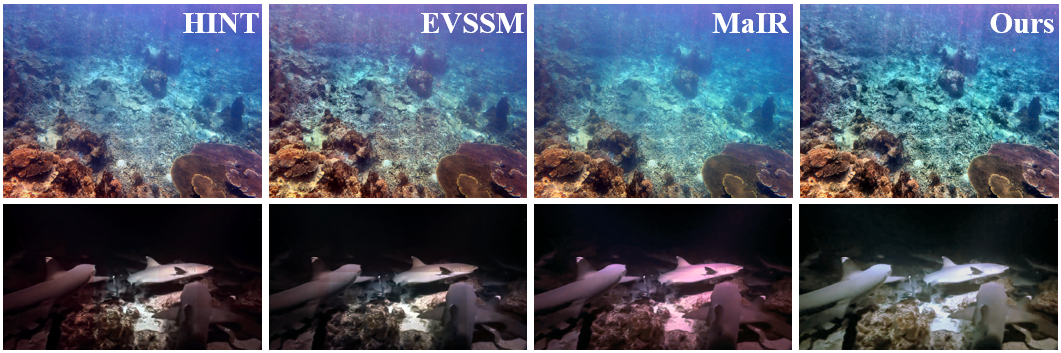}
    \vspace{-6mm}
    \caption{
    \textbf{Full resolution visual comparison.}
}
    \label{fig:full}
    \vspace{-5mm}
\end{figure}

\subsection{Recurrent State Organization Analysis}

We examine whether State Isolation reduces the mixing of hidden representations across tokens with different estimated update tendencies.
Using the $\Delta$-guided routing labels $q_i$, we divide tokens into update-tendency groups and measure how distinguishable their hidden-state features remain after recurrent processing.
Group Separation is defined with $S_{\mathrm{sep}}
=
\frac{V_{\mathrm{between}}}
{V_{\mathrm{between}}+V_{\mathrm{within}}+\epsilon},$
where $V_{\mathrm{between}}$ measures the variation of the group
centers, $V_{\mathrm{within}}$ measures the variation of tokens around
their respective group centers, and $\epsilon$ is a small constant for
numerical stability.
A low value indicates substantial overlap between groups, whereas a high value indicates more distinct group-specific hidden representations.

For Shared State, one Mamba processes the complete token stream through a single recurrent state.
For State Isolation, different $\Delta$-guided groups are processed by dedicated Mamba experts with independent recurrent states.


As shown in Table~\ref{tab:state_separation}, $S_{\mathrm{sep}}$ score increases from 0.1870 to 0.8488 on UIEB and from 0.1906 to 0.8383 on LSUI, with positive improvements for every evaluated image.
These results show that shared state processing produces strongly overlapping representations across update tendency groups, whereas State Isolation yields substantially more distinguishable group-specific representations. 
This quantitatively supports the reduction of cross-group state mixing illustrated in Figure~\ref{fig:m2}. More details are in Suppl (Sec5.1).




\begin{table}[t]
\centering
\caption{
\textbf{Group separation score $S_{\mathrm{sep}}$ of hidden state responses.}
Results are reported as mean$\pm$standard deviation across images.
}
\vspace{-2mm}
\footnotesize
\setlength{\tabcolsep}{9pt}
\begin{tabular}{@{}lccc@{}}
\toprule
Dataset & Shared State & State Isolation & Gain  \\
\midrule
UIEB
& 0.1870$\pm$0.0750
& 0.8488$\pm$0.0512
& $\pm$0.6618 \\

LSUI
& 0.1906$\pm$0.0913
& 0.8383$\pm$0.0586
& $\pm$0.6477 \\
\bottomrule
\end{tabular}
\label{tab:state_separation}
\vspace{-3mm}
\end{table}





\subsection{Ablation Study}
We analyze IsoMamba from coarse-grained framework design to fine-grained internal mechanisms. 
We first evaluate its three top-level components: recurrent-state organization, $\Delta$-derived routing, and channel-adaptive supervision. 
We then examine the internal mechanisms of State Isolation, including instance-adaptive grouping, order-label matching, and global state coordination. 
Finally, we investigate whether the improvement arises from state organization rather than expert capacity and analyze the complementary effect of channel-adaptive supervision on color fidelity.

\begin{figure}[h]
    \centering
    \includegraphics[width=0.98\linewidth]{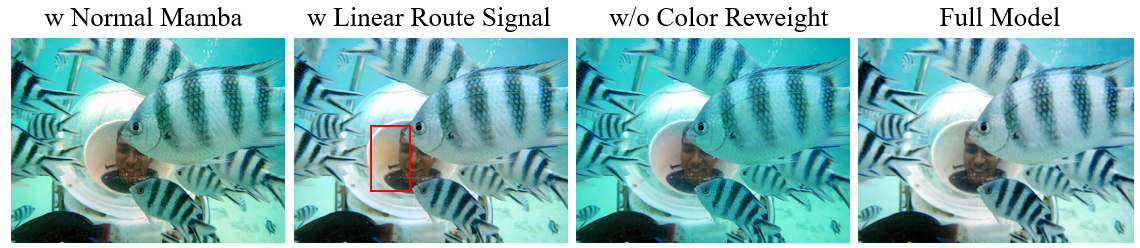}
    \vspace{-2mm}
    \caption{
    \textbf{Qualitative comparison of top-level variants}. Full IsoMamba suppresses the localized yellow artifact around the diver and produces properer foreground colors.}
    \label{fig:abv}
    \vspace{-4mm}
\end{figure}

\begin{figure}[!htbp]
    \centering
    \includegraphics[width=1\linewidth]{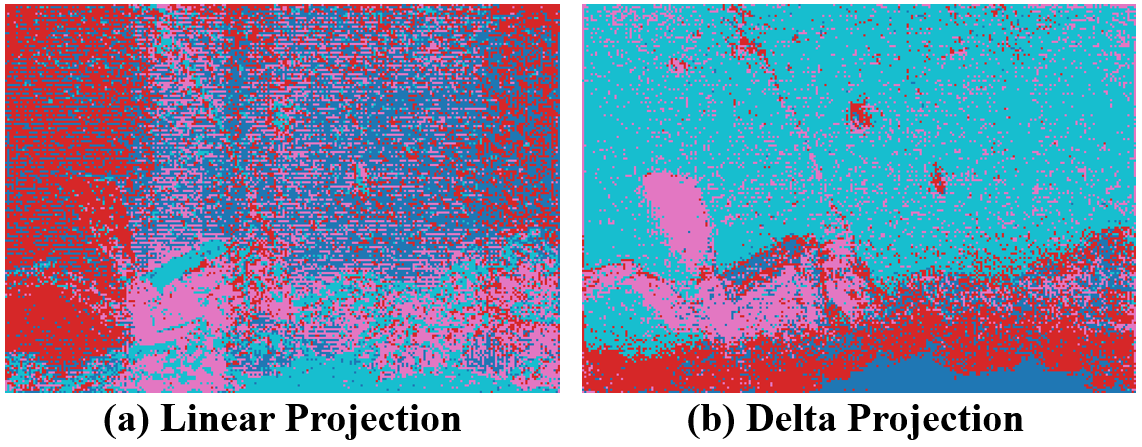}\vspace{-2mm}
   \caption{
    \textbf{Routing assignments from different representations.}
    Colors denote routed expert identities.
    }
    \label{fig:linearP}\vspace{-2mm}
\end{figure}

\textbf{Top-Level Component Ablation.}
Under identical training settings, replacing State Isolation, $\Delta$-derived routing, or channel-adaptive supervision reduces PSNR by 1.28, 0.71, and 1.00 dB, respectively (Table~\ref{tab:ablation-arch}), demonstrating their complementary contributions. Figure~\ref{fig:abv} shows that the full model suppresses the localized yellow artifacts observed around the diver and foreground fish. Figure~\ref{fig:linearP} further explains the routing difference: the generic linear projection produces fragmented expert assignments, whereas the $\Delta$-derived representation forms more spatially coherent groups. Together with the quantitative improvement, this comparison supports $\Delta$ as a more structured, model-intrinsic routing cue. Because the shared-state variant also has fewer parameters, we separately control for expert capacity in Table~\ref{tab:ablation-routing}.

\begin{table}[h]
\centering
\caption{
\textbf{Top-level component ablation.}
}
\vspace{-4mm}
\small
\resizebox{0.48\textwidth}{!}{
\setlength{\tabcolsep}{6pt}
\begin{tabular}{l|ccc}
\toprule
Variant & PSNR $\uparrow$ & SSIM $\uparrow$ & Params (M) \\
\midrule
(i) Shared-state Decoder
& 24.04 & 0.917 & 12.76 \\
(ii) Linear Route Signal
& 24.61 & 0.902 & 15.10 \\
(iii) w/o Channel Reweight
& 24.32 & 0.916 & 15.23 \\
(iv) \textbf{IsoMamba}
& \textbf{25.32} & \textbf{0.920} & 15.23 \\
\bottomrule
\end{tabular}
}
\vspace{-4mm}
\label{tab:ablation-arch}
\end{table}

\begin{figure}[t]
    \centering
    \includegraphics[width=\linewidth]{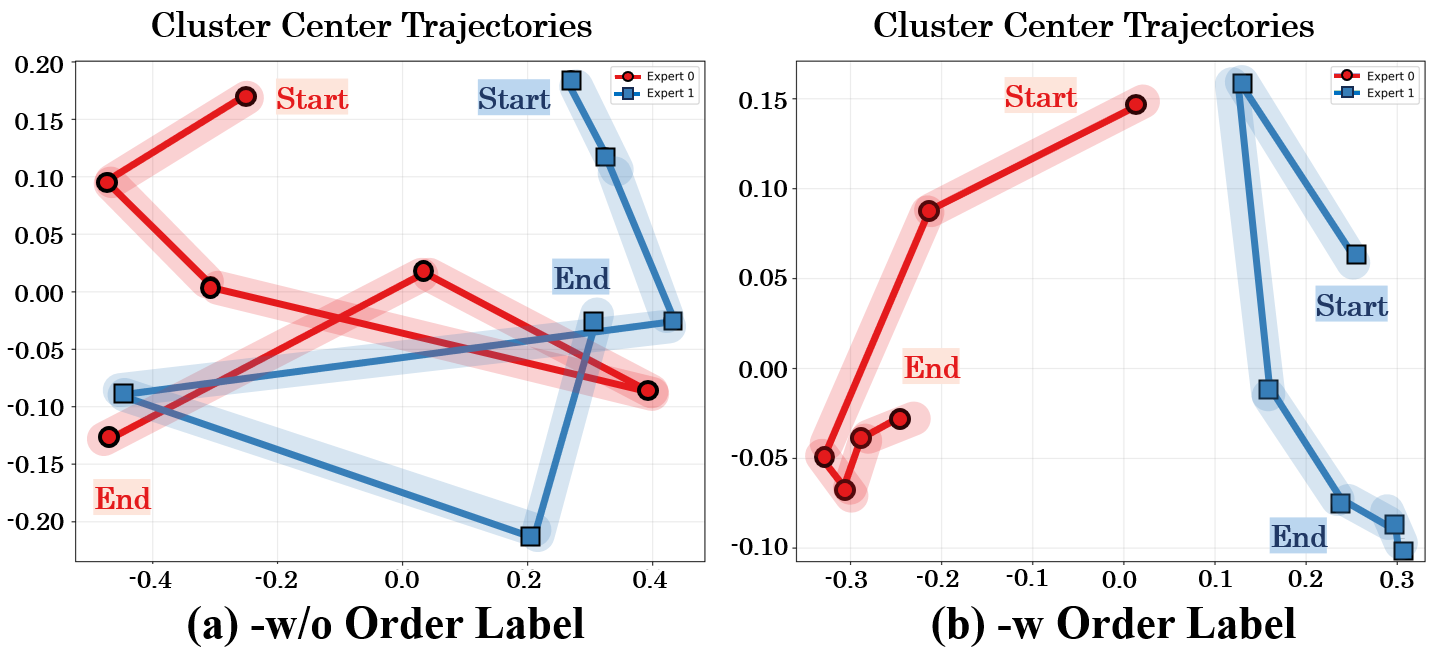}
    \vspace{-2mm}
    \caption{
    \textbf{Effect of order-label matching on cluster--expert correspondence.}
    Ours ordering produces more consistently separated trajectories.
    }
    \label{fig:cluster_center}
\end{figure}

\textbf{Internal Mechanisms of State Isolation.}
Table~\ref{tab:ablation-cluster-internal} examines three internal designs of State Isolation. Replacing instance-adaptive centers with image-invariant prototypes reduces PSNR from 25.32 to 24.51 dB, supporting the benefit of adapting state-sharing groups to each input. Removing order-label matching results in the largest drop to 24.27 dB. As illustrated in Figure~\ref{fig:cluster_center}, unordered centers can switch their associations with expert indices, whereas anchor-based matching yields more consistent cluster--expert trajectories. Finally, removing global state coordination decreases PSNR to 24.88 dB, indicating that isolated recurrent streams still benefit from shared image-level context. Together, these results show complementary roles of adaptive grouping, consistent expert assignment, and global coordination.

\begin{table}[t]
\centering
\caption{
\textbf{Internal mechanisms of State Isolation.}
}
\vspace{-2mm}
\small
\resizebox{0.48\textwidth}{!}{
\begin{tabular}{l|ccc}
\toprule
Variant & PSNR $\uparrow$ & SSIM $\uparrow$ & Params (M) \\
\midrule
(i) Image invariant prototypes
& 24.51 & 0.912 & 14.59 \\
(ii) w/o order label matching
& 24.27 & 0.917 & 15.23 \\
(iii) w/o global state coordination
& 24.88 & 0.918 & 15.06 \\
(iv) \textbf{IsoMamba}
& \textbf{25.32} & \textbf{0.920} & 15.23 \\
\bottomrule
\end{tabular}
}
\label{tab:ablation-cluster-internal}
\vspace{-2mm}
\end{table}

\textbf{State Organization versus Expert Capacity.}
To determine whether the gain arises merely from additional expert capacity, Table~\ref{tab:ablation-routing} compares routing strategies using the same backbone and expert bank.
Soft routing sends the complete token stream through every expert and combines their outputs using learned soft weights.
It therefore retains full expert capacity, but each expert still recurrently processes a heterogeneous token sequence without state isolation, achieving only 24.17~dB.
Top-2 routing similarly mixes responses from multiple recurrent streams and obtains 24.59~dB.
Random Top-1 preserves the same sparse computation as IsoMamba but breaks the correspondence between update tendencies and state streams, resulting in 24.54~dB.
In contrast, $\Delta$-guided Top-1 routing achieves 25.32~dB, showing that the improvement comes from meaningful token-to-state organization rather than expert capacity alone.

\begin{figure}[!htbp]
    \centering
    \includegraphics[width=1\linewidth]{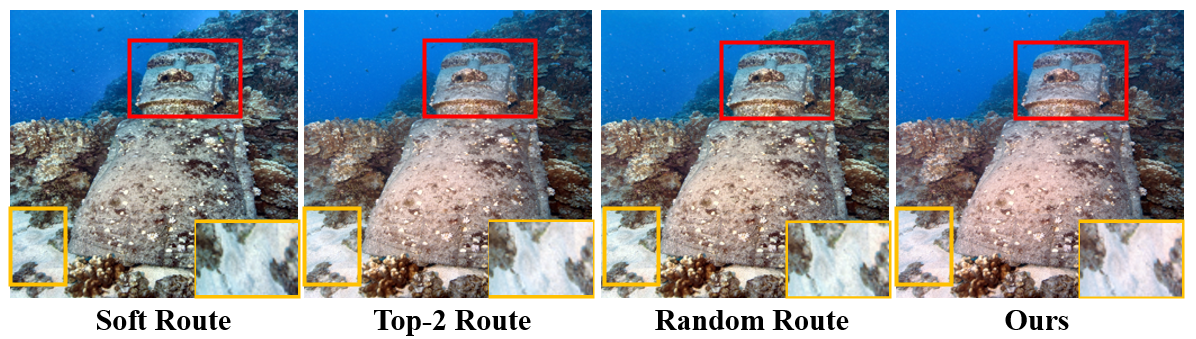}\vspace{-2mm}
    \caption{\textbf{Visual analysis of recurrent-state routing.} Ours routing more effectively corrects the localized blue color cast while introducing fewer noise artifacts in the background.
    }
    \label{fig:vs_RS}
\end{figure}

Figure~\ref{fig:vs_RS} provides complementary visual evidence.
Together, the performance, hidden-state, and visual results support state organization, rather than additional expert capacity, as the main source of improvement.

\begin{table}[!htbp]
\centering
\caption{
\textbf{State organization versus expert capacity.}
}
\vspace{-4mm}
\small
\resizebox{0.48\textwidth}{!}{
\setlength{\tabcolsep}{8pt}
\begin{tabular}{l|ccc}
\toprule
Variant & PSNR $\uparrow$ & SSIM $\uparrow$ & Time (s) \\
\midrule
(i) 
Soft expert mixing   & 24.17 & 0.909 & 0.106 \\
(ii) 
Top-2 expert mixing & 24.59 & 0.912 & 0.102 \\
(iii) Random Top-1 routing                   & 24.54 & 0.913 & 0.089 \\
(iv)  \textbf{$\Delta$-guided Top-1}
& \textbf{25.32} & \textbf{0.920} & 0.095 \\
\bottomrule
\end{tabular}
}
\label{tab:ablation-routing}
\vspace{-2mm}
\end{table}

\section{Conclusion}
This work identifies \emph{State Contention} as a critical limitation of Vision Mamba for spatially heterogeneous restoration, where tokens with incompatible update requirements interfere through a shared recurrent state. We introduce \emph{State Isolation} and instantiate it with IsoMamba, which organizes tokens by model-intrinsic update tendency and routes incompatible groups to independent state streams. Extensive experiments demonstrate that the gains arise from improved recurrent-state organization rather than increased expert capacity. These results highlight recurrent-state organization as a complementary design dimension to token scanning in Vision Mamba.


\clearpage
\onecolumn
\newgeometry{margin=1in}
\setcounter{section}{0}
\setcounter{subsection}{0}
\setcounter{subsubsection}{0}
\setcounter{equation}{0}
\setcounter{figure}{0}
\setcounter{table}{0}
\setcounter{footnote}{0}
\setcounter{page}{1}
\setlength{\floatsep}{6pt}
\setlength{\textfloatsep}{6pt}
\setlength{\intextsep}{6pt}

\null
\vskip 2em
\begin{center}
{\LARGE\bfseries Supplementary Material\par}
\vskip 2mm
{\large State-Isolated Vision Mamba via Update-Tendency Routing
for Underwater Image Enhancement\par}
\end{center}
\vskip 1.5em
\section{Supplementary Material}
Overview. The supplementary includes the following sections:
\vspace{-1mm}
\begin{itemize}
    \item Related Work. (Section \ref{supp:sec:related_work})
    \item High-resolution Full-image Inference. (Section \ref{supp:sec:hr_infer})
    \item Hyperparameter Setting. (Section \ref{supp:sec:hyper_setting})
    \item More Quantitative Analyses. (Section \ref{supp:sec:ema})
    \item More Visualization. (Section \ref{supp:sec:more_vis})
    \item More Applications. (Section \ref{supp:more_app})
    \item Limitations and Future Work. (Section \ref{supp:sec:limitation})
\end{itemize}

\vspace{-4mm}

\section{Related Work}
\label{supp:sec:related_work}
\subsection{Underwater Image Restoration}

Recent underwater image restoration methods increasingly view UIE as a joint global and local modeling problem: wavelength-dependent absorption shifts the overall color distribution, while scattering and depth variation cause spatially varying haze and detail loss \cite{uie-formulation,sea-thru}. Representative designs show a clear architectural trend toward balancing these two needs. PUIE-Net models the ambiguity of paired references with a conditional VAE, where a global latent code modulates local convolutional features, so the model can express multiple plausible enhancement styles \cite{PUIE}. SGUIE-Net pushes adaptivity further by using semantic guidance to apply different enhancement behaviors across regions, but it also introduces reliance on segmentation quality under heavy degradation \cite{SGUIE}. U-shape Transformer brings stronger global interaction through self-attention and combines spatial-wise global modeling with channel-wise fusion for color correction, at the cost of higher compute \cite{lsui}. WF-Diff separates global tone and local details in the frequency domain and uses diffusion to refine high-frequency components, improving perceptual details but adding iterative cost \cite{WD-Diff}. Most recently, SS-UIE adopts Mamba-style selective scanning to obtain global context with linear complexity, aiming at a better speed quality trade-off for high-resolution inputs \cite{ss-uie}.

Overall, these advances improve global consistency and local fidelity, yet the core challenge remains: different clusters within one image can demand incompatible restoration behaviors, which motivates architectures that can adapt their processing more explicitly across space.

\subsection{SSMs for Vision and Image Restoration}
State space models (SSMs) have recently re-emerged as an efficient alternative to attention for long-range sequence modeling. Mamba introduces input-dependent selective updates and achieves linear-time processing, which makes it attractive for high-resolution vision tasks \cite{mamba}. Following this line, Vim and VMamba adapt Mamba to images by serializing 2D features into scan-ordered sequences with multi-directional scans so that each token can access global context with linear complexity \cite{visionMamba,vmamba}. Building on these visual backbones, restoration-oriented designs such as MambaIR further integrate SSM blocks into image restoration pipelines and augment the vanilla scan-based module with local enhancement and channel attention to better preserve textures and handle channel redundancy \cite{MambaIR}.

To better cope with diverse degradations, recent works start to combine Mamba with mixture-of-experts (MoE) routing. Uni-UIR introduces expert specialization for distinct underwater degradations and uses a gating module to route features to the most suitable experts \cite{uni-uir}. M2-Restore further proposes a CLIP-guided MoE router that fuses task-conditioned prompts with CLIP-derived semantic priors, and refines routing via cross-modal feature calibration to select experts for different degradation types \cite{m2restore}. Beyond routing, several studies also begin to revisit structural limitations of scan-based SSMs. MambaIRv2 equips SSM restoration with non-causal interactions, moving closer to the bidirectional information mixing of vision transformers \cite{MambaIRv2}. Spatial-Mamba highlights the spatial-topology mismatch introduced by 2D-to-1D serialization, and proposes neighborhood state fusion in the state space to better preserve spatial structure even under a single scan \cite{spatialmamba}. 

These developments suggest that improving visual SSM restoration is not only about stronger routing cues, but also about addressing structural issues induced by serialization and state evolution, which motivates our design.

\section{High-resolution Full-image Inference}
\label{supp:sec:hr_infer}

All main-table benchmark numbers follow the common $256\times256$ resize protocol for fair comparison under limited GPU memory in the main paper.
Here we additionally report full-image inference on the original high resolution without tiling.
We use a single RTX4090 24GB GPU with batch size 1.
For each image, we apply reflect padding to make the spatial size divisible by 8, run one forward pass, then remove the padding.
We record runtime per image and the peak GPU memory, measured as the maximum CUDA memory allocated.
We also report the maximum CUDA memory reserved to reflect the caching allocator behavior.

Table~\ref{supp:tab:hr_top5} summarises the five largest test images across existing common datasets (UIEB, LSUI, EUVP).
Our model supports direct inference up to $1800\times1350$, 2.43 MP, with 3.49 seconds runtime and 17.37 GB peak allocated memory on the largest sample.
Across the top five largest images, the peak allocated memory stays within 16.64 to 17.37 GB, while producing reasonable PSNR and SSIM at the original resolution.

\begin{table}[htbp]
\centering
\caption{Full-image inference on the five largest UIEB test images using one RTX 4090 GPU (batch size 1).}
\label{supp:tab:hr_top5}
\footnotesize
\setlength{\tabcolsep}{2pt}
\resizebox{\columnwidth}{!}{%
\begin{tabular}{@{}lcccc@{}}
\toprule
Image & Resolution & Time (s) & Alloc. (GB) & PSNR / SSIM \\
\midrule
689\_img\_.png & $1800\times1350$ & 3.486 & 17.37 & 24.28 / 0.959 \\
720\_img\_.png & $1800\times1300$ & 3.181 & 16.74 & 18.41 / 0.813 \\
731\_img\_.png & $1800\times1295$ & 3.162 & 16.67 & 22.93 / 0.945 \\
772\_img\_.png & $1800\times1295$ & 3.192 & 16.66 & 24.33 / 0.895 \\
796\_img\_.png & $1800\times1295$ & 3.202 & 16.64 & 19.83 / 0.836 \\
\bottomrule
\end{tabular}%
}
\end{table}


\begin{figure}[!htbp]
        \centering
        \vspace{-1mm}
        \includegraphics[width=0.98\linewidth]{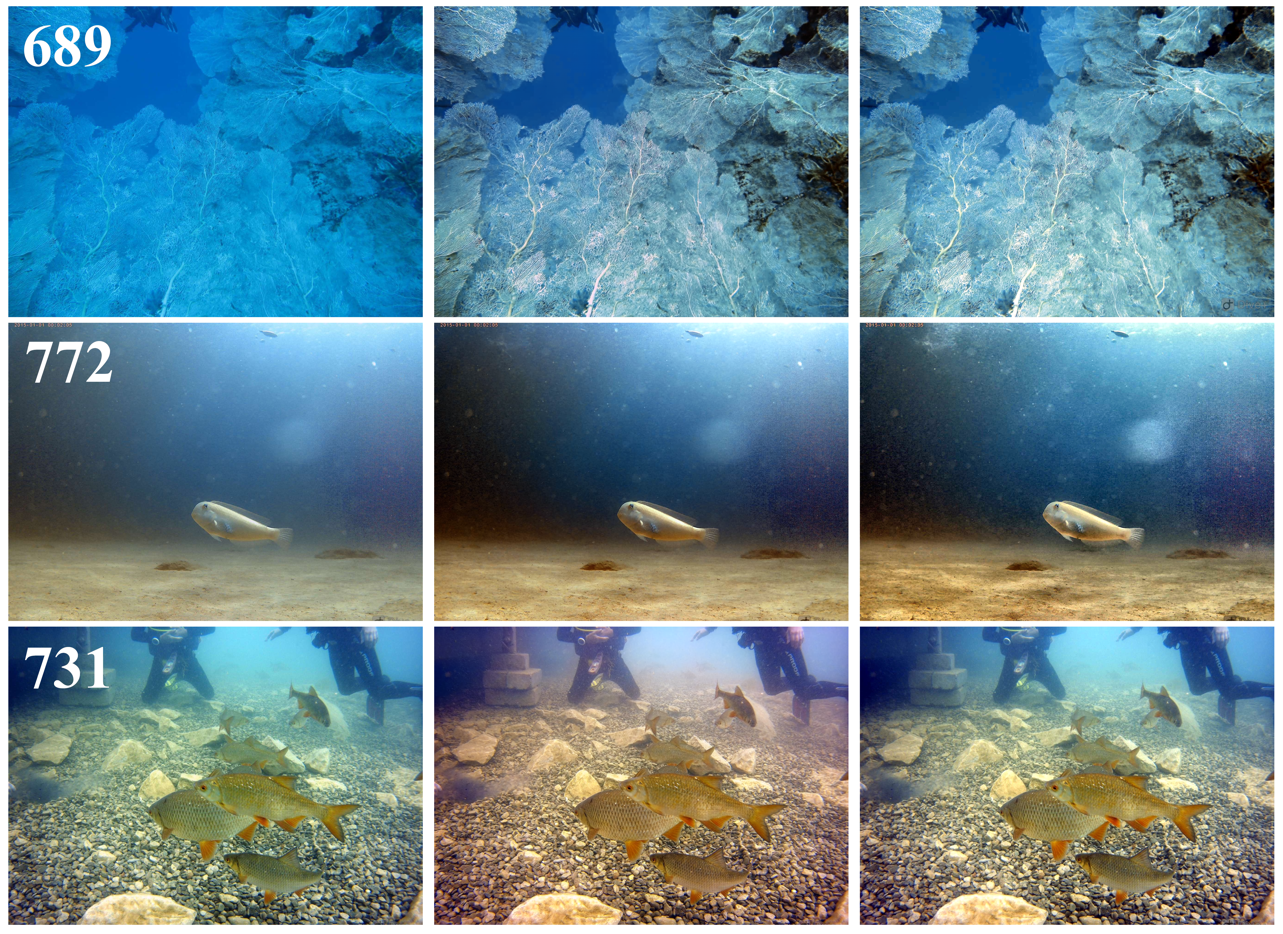}
        \caption{Qualitative full-image results on large-resolution test images.
        Each row shows one sample, indexed by the file id.
        From left to right are the input, our output, and the ground truth.
        All results are produced by a single full-image forward pass without tiling.}
        \vspace{-3mm}
        \label{supp:fig:Visual Comparison on high resolution}
\end{figure}

\section{Hyperparameter Setting}
\label{supp:sec:hyper_setting}
We use a four-stage encoder--decoder with base channel dimension 32 and block numbers $[4,6,6,8]$.
IsoMamba maintains a small set of experts per decoder stage, controlled by \texttt{num\_clusters}, which determines the number of cluster centers and thus the number of disjoint spatial groups routed to independent state streams at that stage.

Our default setting uses \texttt{num\_clusters} $=[2,3,4]$, allocating more groups to the coarsest-scale decoder stage (near the bottleneck), where multi-scale features are most mixed and tokens tend to exhibit more diverse update tendencies.
We also validated alternative schedules, including a reversed allocation $[4,3,2]$ and a uniform allocation $[3,3,3]$.
We report the comparison under the same training and evaluation protocol to study the sensitivity to expert count and its distribution across stages.

\begin{table}[htbp]
\centering
\caption{Effect of the number of clusters.}
\small
\setlength{\tabcolsep}{3pt}
\begin{tabular}{l|cccc}
\toprule
Variant  & PSNR  & SSIM  & Param (M)& Time (s) \\
\midrule
\texttt{-w/} [3,3,3]    & 25.28  &  0.917  &14.62 & 0.093 \\
\texttt{-w/} [4,3,2]     & 24.87  &  0.919  & 14.02  & 0.090\\
\texttt{-w/} [2,3,4] (Ours) & 25.32  &  0.920 & 15.23  &  0.097\\
\bottomrule
\end{tabular}
\label{supp:tab:ema-orderlabel-1}
\end{table}

At the same time, we provide the hyperparameter details and computational complexity for per-batch clustering here.
Given a pooled feature map $\mathrm{pooled}\in\mathbb{R}^{B\times H\times W\times R}$, we reshape it into token form
$X\in\mathbb{R}^{B\times N\times R}$ with $N=HW$, and apply $\ell_2$ normalization along the feature dimension.
We then compute $K$ cluster centers for each batch element using a k-means++ style initialization followed by $T$ iterations of assignment-and-update.

\paragraph{Complexity of initialization.}
Starting from one uniformly sampled token per batch, we adds one center at a time.
At step $m$, it computes the squared distance from each token to its nearest selected center among $m-1$ centers.
A straightforward implementation therefore costs
\[
\mathcal{O}\!\left(B\cdot N\cdot R\cdot K\right),
\]
where the $\mathcal{O}(B\cdot N\cdot R)$ term for sampling the first center is negligible compared with the above.

\paragraph{Complexity of iterative refinement.}
Each iteration computes the full token-to-center distance matrix in $\mathbb{R}^{B\times N\times K}$, followed by an hard assignment, and then updates centers via weighted averaging.
The dominant costs are (i) distance computation and (ii) weighted aggregation, both of which scale as $\mathcal{O}(B\cdot N\cdot K\cdot R)$.
Thus, with $T$ refinement iterations, the overall refinement complexity is
\[
\mathcal{O}\!\left(B\cdot T\cdot N\cdot K\cdot R\right).
\]

\paragraph{Total complexity and the $16\times$ reduction.}
Combining initialization and refinement, the total clustering cost is
\[
\mathcal{O}\!\left(B\cdot N\cdot R\cdot K\right)\;+\;\mathcal{O}\!\left(B\cdot T\cdot N\cdot K\cdot R\right) \\
\;\approx\;
\mathcal{O}\!\left(B\cdot (T+1)\cdot N\cdot K\cdot R\right).
\]
Let $(N_0, R_0)$ denote the token count and feature dimension at the original input resolution.
In practice, we run the clustering on reduced features: we downsample the spatial resolution by $2\times 2$ so that $N$ becomes $4\times$ smaller than $N_0$, and we also reduce the channel dimension from the input to $R=\frac{1}{4}R_0$.
Therefore, we perform clustering on reduced features with $N = \frac{N_0}{4}$ and $R=\frac{R_0}{4}$, hence the dominant term
$B\cdot (T+1)\cdot N\cdot K\cdot R$ is reduced by
\[
\frac{N}{N_0}\cdot \frac{R}{R_0} \;=\; \frac{1}{4}\cdot\frac{1}{4}\;=\;\frac{1}{16}.
\]
This $16\times$ cost reduction makes the clustering overhead small enough to keep the overall inference time within sub-second latency in our implementation.

\section{More Quantitative Analyses.}
\label{supp:sec:ema}

\subsection{Quantifying Raw Hidden-State Separation}

We directly examine whether State Isolation changes the organization of the
recurrent states themselves, rather than only affecting their subsequent
readout. Specifically, we analyze the raw selective-scan state
$\mathbf{h}_i^{(r)}$ recorded immediately after updating token $i$ in scan
direction $r$.The analysis
isolates the recurrent state before any readout or output transformation.

\paragraph{Fixed-expert control.}
To distinguish the effect of state isolation from that of expert
specialization, we construct two fixed-expert response modes from the same
trained checkpoint. In \emph{Shared State}, all tokens are processed using
expert~0 and follow one continuous recurrent trajectory per scan direction. In
\emph{Isolated State}, the same tokens are partitioned using the model's actual
$\Delta$-based routing assignments and processed through independent
group-wise recurrent trajectories, while all groups still use expert~0.
The two modes reuse identical token-wise operands, including
$\mathbf{x}_i$, $\Delta_i$, $B_i$, $A$, and the delta bias. Consequently, their
only difference is the recurrence organization: a token in the shared mode
inherits the state accumulated along the original scan, whereas a token in the
isolated mode inherits only the state history of its routed group. Dummy
padding positions are excluded from both recurrence and measurement.

We additionally report \emph{Full Isolation}, in which each routed group uses
its corresponding expert and maintains an independent recurrent trajectory.
This setting describes the combined effect of state isolation and expert
specialization; it is not used to identify the state-isolation effect.

\paragraph{Separation metric.}
For each of the four scan directions, we map the raw state trajectory back to
the original spatial token positions without merging directions. We flatten
and normalize the state associated with token $i$ as
\[
\widehat{\mathbf{h}}_i^{(r)}
=
\frac{
\operatorname{vec}\!\left(\mathbf{h}_i^{(r)}\right)
}{
\left\|
\operatorname{vec}\!\left(\mathbf{h}_i^{(r)}\right)
\right\|_2+\epsilon
},
\]
where $\epsilon$ is a small constant for numerical stability. Let
$\mathcal{G}_k$ denote the set of tokens assigned to routing group $k$,
$N$ the total number of tokens, and $K$ the number of active groups. The
group-specific and global state centers for direction $r$ are defined as
\[
\boldsymbol{\mu}_k^{(r)}
=
\frac{1}{|\mathcal{G}_k|}
\sum_{i\in\mathcal{G}_k}
\widehat{\mathbf{h}}_i^{(r)},
\qquad
\boldsymbol{\mu}^{(r)}
=
\frac{1}{N}
\sum_{i=1}^{N}
\widehat{\mathbf{h}}_i^{(r)}.
\]
We then compute the between-group and within-group state dispersions:
\[
V_{\mathrm{between}}^{(r)}
=
\sum_{k=1}^{K}
|\mathcal{G}_k|
\left\|
\boldsymbol{\mu}_k^{(r)}
-
\boldsymbol{\mu}^{(r)}
\right\|_2^2,
\]
\[
V_{\mathrm{within}}^{(r)}
=
\sum_{k=1}^{K}
\sum_{i\in\mathcal{G}_k}
\left\|
\widehat{\mathbf{h}}_i^{(r)}
-
\boldsymbol{\mu}_k^{(r)}
\right\|_2^2.
\]
The directional hidden-state separation is defined as
\[
S_{\mathrm{state}}^{(r)}
=
\frac{
V_{\mathrm{between}}^{(r)}
}{
V_{\mathrm{between}}^{(r)}
+
V_{\mathrm{within}}^{(r)}
+\epsilon
}.
\]
A higher value indicates that raw recurrent states are more compact within
each routed group and more distinct across groups. Since the four scan
directions maintain independent recurrent trajectories, we compute the score
separately for each direction and report their arithmetic mean:
\[
S_{\mathrm{state}}
=
\frac{1}{4}
\sum_{r=1}^{4}
S_{\mathrm{state}}^{(r)}.
\]
The score is first computed for each test image and then aggregated over the
dataset.

\paragraph{Analysis.}
As shown in Table~\ref{supp:tab:raw_state_separation}, group-wise recurrence
isolation substantially increases the separation of raw hidden states even
when the expert parameters and all token-wise recurrence operands are held
fixed. On UIEB, the score increases from $0.1870$ to $0.5765$, corresponding
to a paired improvement of $0.3895$ with a 95\% confidence interval of
$[0.3781,0.4007]$. On LSUI, it increases from $0.1906$ to $0.5614$, with a
paired improvement of $0.3707$ and a 95\% confidence interval of
$[0.3643,0.3770]$. The improvements are significant under both paired
$t$-tests and Wilcoxon signed-rank tests on each dataset
(all $p<2\times10^{-16}$), with Cohen's $d_z$ values of $6.99$ on UIEB and
$5.73$ on LSUI.

The similar shared-state scores, isolated-state scores, and paired
improvements across the two datasets indicate that the effect is consistent
rather than dataset-specific. Because the fixed-expert comparison removes
differences in expert parameters and excludes all state readout and output
transformations, it provides direct evidence that group-wise recurrence
isolation changes the structure of the raw recurrent states themselves.
In particular, restricting the recurrent history to tokens from the same
$\Delta$-defined group produces states that are more compact within groups
and more distinct across groups.

Full Isolation further increases the score to $0.8488$ on UIEB and $0.8383$
on LSUI. This additional increase suggests that expert specialization
amplifies the separated state structure established by group-wise recurrence
isolation. However, since Full Isolation also changes the expert parameters,
its difference from Shared State should not be interpreted as a pure
state-isolation effect.


Moreover,
the routing assignments define the groups used by the separation metric.
Accordingly, the analysis demonstrates that State Isolation makes raw
recurrent states conform more strongly to the learned $\Delta$-based routing
structure, but does not by itself establish that these groups are uniquely
optimal or semantically defined.

\begin{table}[t]
\centering
\caption{
\textbf{Separation of raw recurrent hidden states.}
Shared and Isolated State use identical expert-0 operands; Full Isolation
uses the corresponding expert for each group. Scores are mean$\pm$standard
deviation. The paired gap is Isolated minus Shared State.
}
\label{supp:tab:raw_state_separation}
\footnotesize
\setlength{\tabcolsep}{3pt}
\begin{tabular}{@{}lcc@{}}
\toprule
Measure & UIEB & LSUI \\
\midrule
Shared State (fixed expert)
  & $0.1870\pm0.0750$ & $0.1906\pm0.0913$ \\
Isolated State (fixed expert)
  & $0.5765\pm0.0933$ & $0.5614\pm0.1019$ \\
Full Isolation
  & $0.8488\pm0.0512$ & $0.8383\pm0.0586$ \\
\midrule
Paired gap
  & $0.3895$ & $0.3707$ \\
95\% bootstrap CI
  & $[0.3781,\,0.4007]$ & $[0.3643,\,0.3770]$ \\
\bottomrule
\end{tabular}
\end{table}



\subsection{Channel-Adaptive Supervision}

Complementary to spatial State Isolation, channel-adaptive reweighting
accounts for the unequal reconstruction difficulty of RGB channels in
underwater scenes. Removing this supervision decreases PSNR from
25.32 to 24.32 dB. As shown in Table~\ref{supp:tab:kl-color}, it also increases
the average channel-wise color-histogram KL divergence from 0.0064 to
0.0071. The largest improvement is observed in the green channel
($0.0051\rightarrow0.0036$), while the easiest blue channel exhibits only a
minor increase ($0.0038\rightarrow0.0041$). Overall, these results show
that channel adaptive supervision improves color fidelity while providing
a substantial gain in reconstruction accuracy.

\begin{table}[h]
\centering
\caption{
\textbf{Channel-wise color-histogram KL divergence.}
Lower values indicate better agreement with the ground-truth color
distribution.
}
\label{supp:tab:kl-color}
\small
\setlength{\tabcolsep}{3pt}
\begin{tabular}{@{}lcccc@{}}
\toprule
Variant & Red & Green & Blue & Avg. $\downarrow$ \\
\midrule
Degraded Input
& 0.1574 & 0.0712 & 0.0608 & 0.0965 \\

\texttt{w/o} Channel Reweighting
& 0.0125 & 0.0051 & \textbf{0.0038} & 0.0071 \\

\textbf{IsoMamba(Ours)}
& \textbf{0.0117} & \textbf{0.0036} & 0.0041 & \textbf{0.0064} \\
\bottomrule
\end{tabular}
\end{table}

\subsection{Contribution of Loss Functions.}
Table~\ref{supp:tab:loss_ablation} investigates the contribution of each loss component.
Removing the $\ell_1$ loss decreases the PSNR from 25.32 dB to 24.76 dB, indicating that pixel-level supervision is important for preserving reconstruction fidelity.
Without the SSIM loss, the performance drops significantly to 23.76 dB PSNR, suggesting that structural consistency plays an important role in maintaining perceptual quality.
Removing the edge loss also leads to degraded performance in both PSNR and UCIQE, demonstrating that edge-aware constraints help recover high-frequency details and improve visual quality.
Overall, the full objective function achieves the best performance across all evaluation metrics.

\begin{table}[h]
\centering
\caption{
\textbf{Ablation study on loss function components.}
}
\vspace{-2mm}
\footnotesize
\setlength{\tabcolsep}{6pt}
\begin{tabular}{l|ccc}
\toprule
Variant & PSNR$\uparrow$ & SSIM$\uparrow$ & UCIQE$\uparrow$ \\
\midrule
(i) w/o $\ell_1$ Loss 
& 24.76 & 0.916 & 0.617 \\
(ii) w/o SSIM Loss
& 23.76 & 0.902 & 0.615 \\
(iii) w/o Edge Loss
& 24.52 & 0.909 & 0.612 \\
\textbf{IsoMamba (Ours)}
& \textbf{25.32} & \textbf{0.920} & \textbf{0.618} \\
\bottomrule
\end{tabular}
\label{supp:tab:loss_ablation}
\end{table}

\subsection{EMA Discussion}
We additionally test an EMA-based update for the order label, where the label is updated during training by an exponential moving average of the sorted cluster centers.
This adaptive ordering slightly improves performance on LSUI (29.89/0.914 $\rightarrow$ 29.97/0.915), but reduces generalization on UIEB (25.32/0.920 $\rightarrow$ 24.81/0.917).
A likely reason is data imbalance: LSUI contributes about 75\% of the training images, so the EMA label is biased toward LSUI statistics and becomes less suitable for UIEB.
Given this train-domain gain but cross-domain drop, we keep the fixed order label in all experiments for robustness.

\begin{table}[htbp]
\centering
\caption{Effect of EMA-updated order label on different test sets.}
\small
\setlength{\tabcolsep}{4pt}
\begin{tabular}{l|cc}
\toprule
Variant  & UIEB  & LSUI  \\
\midrule
\texttt{-w/} EMA-updated  & 24.81 / 0.917  & 29.97 / 0.915 \\
\texttt{-w/o} EMA-updated (Ours)    & 25.32 / 0.920  & 29.89 / 0.914 \\
\bottomrule
\end{tabular}
\label{supp:tab:ema-orderlabel-2}
\end{table}

\section{More Visualization}
\label{supp:sec:more_vis}
\begin{figure}[!h]
    \centering
    \includegraphics[width=0.98\linewidth]{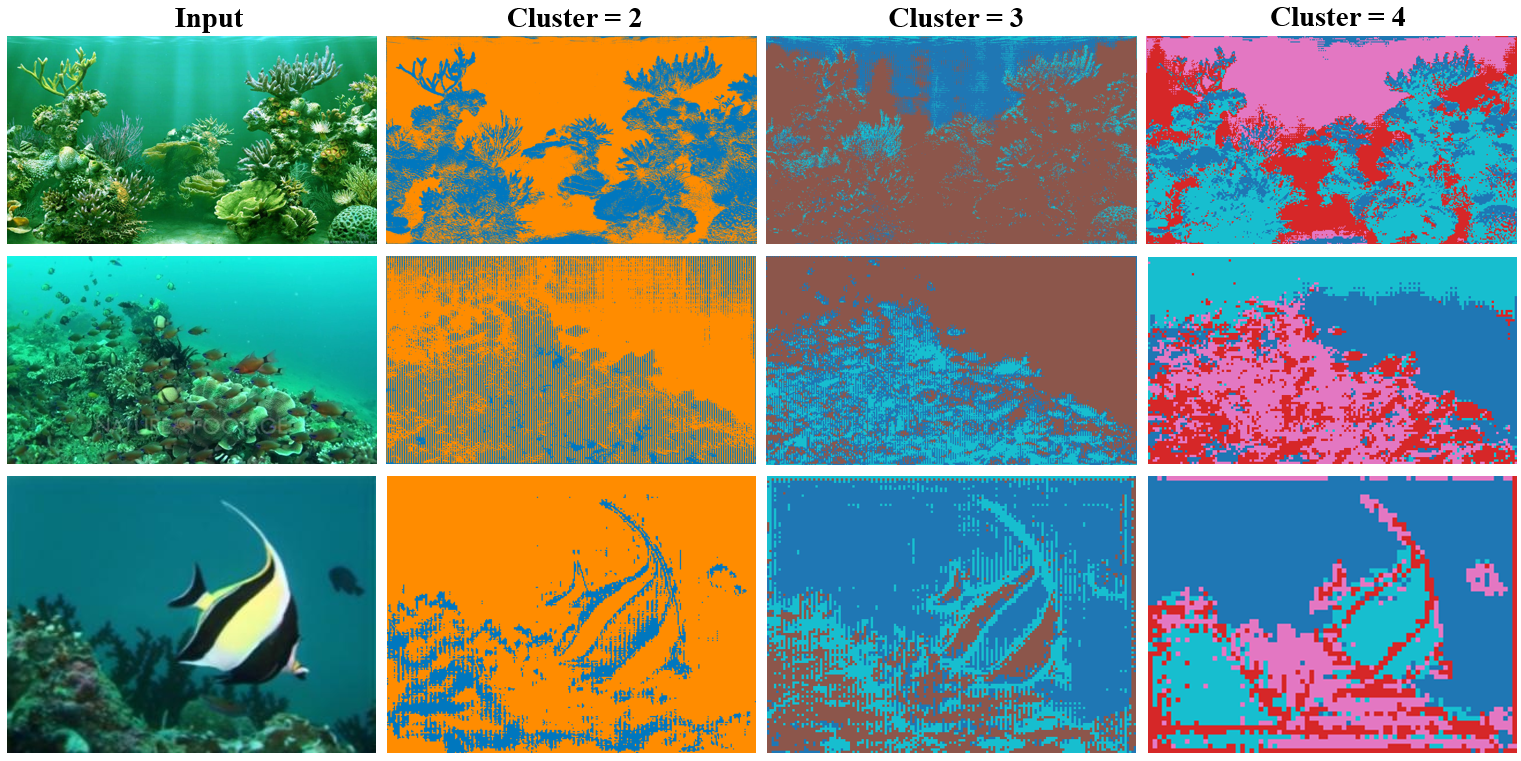}
    \caption{
    Visualization of $\Delta$-driven cluster maps with different numbers of clusters.
    From left to right, we show the input image and the route maps for $K=2$, $K=3$, and $K=4$.
    Each color denotes one cluster and its routed expert identity.
    }
    \label{supp:fig:cluster_map}
\end{figure}

\paragraph{Visualization of routing partitions under different numbers of clusters.}
We visualize the cluster route maps produced by our $\Delta$-based routing under different cluster counts $K$ from different stages in Figure ~\ref{supp:fig:cluster_map}.
For each input image, we compute the update-tendency descriptor derived from $\Delta$, perform spatial clustering with $K \in \{2,3,4\}$, and render the resulting per-pixel cluster identity as a color-coded map.
Even without any semantic labels, the partitions are spatially coherent and often separate large smooth water clusters from more textured structures.
As $K$ increases, complex areas are further subdivided into finer groups, which is consistent with the idea that different clusters may prefer different state update behaviors and therefore benefit from being routed to different state streams.

\begin{figure}[!h]
    \centering
    \includegraphics[width=0.98\linewidth]{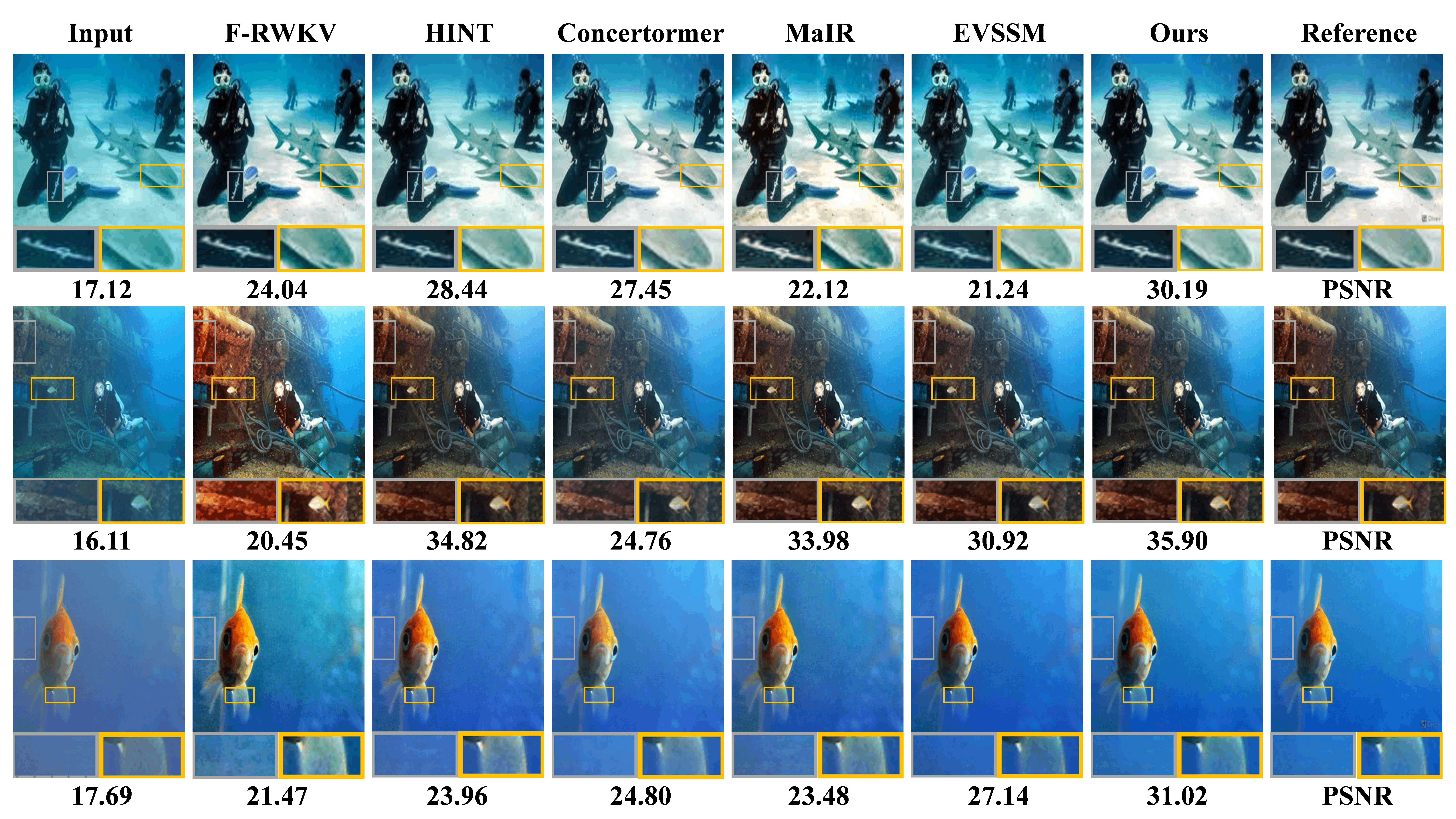}
    \caption{
    Additional qualitative comparisons on representative test images.
    We compare with F-RWKV, HINT, Concertormer, MaIR, and EVSSM.
    Grey and yellow boxes indicate zoom-in regions, and the number under each result is PSNR for that image.
    }
    \label{supp:fig:v1}
\end{figure}

\paragraph{Additional visual comparisons.}
We provide additional visualisations for both full-reference datasets and non-reference datasets. Figure~\ref{supp:fig:v1} provides qualitative comparisons against representative baselines on full-reference benchmarks. We include zoomed-in patches to emphasize challenging regions and report the PSNR for each example. As shown in Figure~\ref{supp:fig:v1}, our method consistently yields more accurate colour correction, recovering the underlying hues in the first two rows while maintaining appropriate saturation (for example, the yellow tail of the small fish). Moreover, for the blurred case in the third row, our method restores the fine structures of the fish fins more completely, and avoids amplifying noise in visually smooth water regions that can arise from overly aggressive enhancement.

\begin{figure}[!h]
    \centering
    \includegraphics[width=0.98\linewidth]{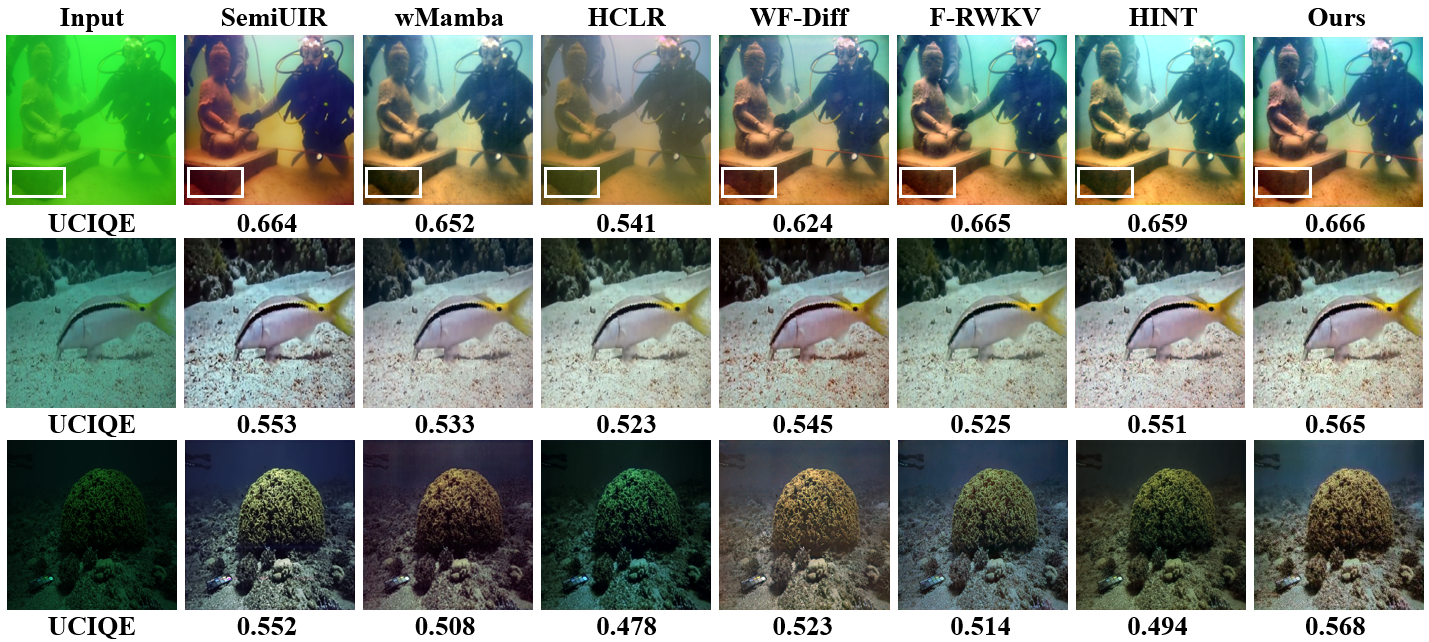}
    \caption{
    Additional NR visual comparisons on UCIQE.
    We compare with representative strong baselines under this metric, and report the per-image UCIQE score beneath each result.
    }
    \label{supp:fig:v2}
\end{figure}

Moreover, Figure~\ref{supp:fig:v2} presents qualitative comparisons on no-reference  benchmarks, where ground-truth images are unavailable. We include representative methods with strong UCIQE performance, and report the per-image UCIQE score beneath each output. As shown in Figure~\ref{supp:fig:v2}, we evaluate three challenging cases: severe green cast, blue-green cast, and low-light conditions with increasing water depth. In the first row, our method improves sharpness while suppressing noise within the highlighted regions. In the second row, the restoration on sandy terrain remains colour-balanced, avoiding noticeable over-compensation toward blue or red tones. In the third row, our method recovers brightness more effectively and produces clearer structures under low illumination.

\section{More Applications}
\label{supp:more_app}
In addition to the underwater dataset, we also conduct \textbf{single-task experiments} on conventional restoration tasks, including \textbf{denoising, dehazing, deraining, deblurring, and low-light enhancement}. For each task, we train a separate model using the corresponding training set and evaluate it on the matching test set. The dataset details and the train test pairing are summarized in Table~\ref{supp:tab:dataset}.

\begin{table}[htbp]
\centering
\caption{Single-task dataset settings for conventional restoration.
Each task is trained and tested independently; parentheses show image counts.}
\label{supp:tab:dataset}
\footnotesize
\setlength{\tabcolsep}{4pt}
\begin{tabular}{@{}llp{0.58\columnwidth}@{}}
\toprule
\rowcolor{blue!5}
\textbf{Task} & \textbf{Split} & \textbf{Dataset (images)} \\
\midrule
\multirow{2}{*}{\textcolor{red}{Noise}}
  & Train & BSD400~\cite{BSD400} + WED~\cite{WED} (400 + 4,744) \\
  & Test  & Kodak24~\cite{Kodak24} (24) \\
\addlinespace
\multirow{2}{*}{\textcolor{blue}{Haze}}
  & Train & RESIDE-$\beta$-OTS~\cite{beta-OTS} (72,135) \\
  & Test  & SOTS-Outdoor~\cite{SOTS-outdoor} (500) \\
\addlinespace
\multirow{2}{*}{\textcolor{green!60!black}{Rain}}
  & Train & Rain100L~\cite{Rain100L} (200) \\
  & Test  & Rain100L~\cite{Rain100L} (100) \\
\addlinespace
\multirow{2}{*}{\textcolor{purple}{Blur}}
  & Train & GoPro~\cite{Gopro} (2,103) \\
  & Test  & GoPro~\cite{Gopro} (1,111) \\
\addlinespace
\multirow{2}{*}{\textcolor{pink!70!black}{Low-Light}}
  & Train & LOL~\cite{LOL} (485) \\
  & Test  & LOL~\cite{LOL} (15) \\
\bottomrule
\end{tabular}
\end{table}



For \textcolor{red}{\textbf{denoising}}, we include DnCNN~\cite{DNCNN}, FFDNet~\cite{FFDNet}, ADFNet~\cite{ADFNet}, DGUNet~\cite{DGUNet}, Restormer~\cite{Restormer}, and FSNet~\cite{FSNet}. 
For \textcolor{blue}{\textbf{dehazing}}, we evaluate DehazeNet~\cite{DehazeNet}, FDGAN~\cite{FD-GAN}, DehazeFormer~\cite{DehazeFormer}, Restormer~\cite{Restormer}, NAFNet~\cite{NAFnet}, and FSNet~\cite{FSNet}. 
For \textcolor{green}{\textbf{deraining}}, we compare with MSPFN~\cite{MSPFN}, LPNet~\cite{LPNet}, DRSformer~\cite{DRSformer}, Restormer~\cite{Restormer}, and FSNet~\cite{FSNet}. 
For \textcolor{purple}{\textbf{deblurring}}, we adopt DeblurGAN~\cite{DeblurGAN}, Stripformer~\cite{Stripformer}, HI-Diff~\cite{HI-Diff}, MPRNet~\cite{MPRNet}, Restormer~\cite{Restormer}, and FSNet~\cite{FSNet}. 
For \textcolor{pink}{\textbf{low-light enhancement}}, we include URetinex~\cite{URetinex}, SMG~\cite{SMG}, Retinexformer~\cite{Retinexformer}, Restormer~\cite{Restormer}, and DiffIR~\cite{DiffIR}. 

We also report representative general-purpose and \textbf{All-In-One} methods, including AirNet~\cite{AirNet}, IDR~\cite{IDR}, PromptIR~\cite{PromptIR}, and Perceive-IR~\cite{Perceive-IR}, trained under the same single-task protocol for a fair comparison.

To verify that our delta-aware design is not overfitted to underwater domains, we evaluate IsoMamba on five standard restoration benchmarks under a single-task setting, as shown in Table~\ref{supp:tab:5t}.

(1) \textcolor{red}{\textbf{Denoising}}: On Kodak24, IsoMamba achieves the best PSNR across all noise levels. Notably, at the heavy-noise setting ($\sigma=50$), it outperforms the runner-up by a clear margin (29.58 vs. 29.16 dB), demonstrating robust feature preservation under severe corruption. (2) \textcolor{blue}{\textbf{Dehazing}}: On SOTS, our method achieves a remarkable breakthrough, surpassing the second-best method by over 1.0 dB in PSNR (32.80 vs. 31.78 dB). This confirms that our explicit handling of state dynamics is particularly effective for atmospheric degradation. (3) \textcolor{green}{\textbf{Deraining}} and  \textcolor{pink}{\textbf{Low-light}}: On Rain100L, we achieve the best SSIM and the second-best PSNR. On LOL, IsoMamba ranks in the top three, outperforming other generalist baselines like Perceive-IR and AirNet, falling short only to specialized low-light models (e.g., Retinexformer). (4) \textcolor{purple}{\textbf{Deblurring}}: On GoPro, while not achieving state-of-the-art performance, our model maintains competitive ability (31.45 dB), indicating reasonable generalization capability even without motion-specific architectural priors.

These results indicate that our state-isolation design complements strong general restoration priors by clustering the gate ($\Delta$) signal, leading to a balanced improvement profile rather than targeting a single task.

\begin{table*}[ht]
\centering
\small
\setlength{\tabcolsep}{4.5pt}
\renewcommand{\arraystretch}{1.12}
\caption{Single-task restoration results, including denoising on Kodak24, dehazing on SOTS, deraining on Rain100L, deblurring on GoPro, and low-light enhancement on LOL. The top three PSNR results are marked in \textcolor{best}{RED}, \textcolor{second}{BLUE}, and \textcolor{third}{GREEN}, respectively.}
\label{supp:tab:single_task_results}
\resizebox{\textwidth}{!}{
\begin{tabular}{lccc|lc|lc|lc|lc}
\toprule
\cellcolor{blue!5}Kodak24 & \multicolumn{3}{c|}{PSNR} & \cellcolor{blue!5}SOTS & PSNR/SSIM & \cellcolor{blue!5}Rain100L & PSNR/SSIM & \cellcolor{blue!5}GoPro & PSNR/SSIM & \cellcolor{blue!5}LOL & PSNR/SSIM \\
\cmidrule(lr){2-4}
 & $\sigma = 15$ & $\sigma = 25$ & $\sigma = 50$ &  &  &  &  &  &  &  &  \\
\midrule
DnCNN      & 34.60 & 32.14 & 28.95 & DehazeNet      & 22.46/0.851 & MSPFN        & 33.50/0.948 & DeblurGAN   & 28.70/0.858 & URetinex      & 21.33/0.835 \\
FFDNet     & 34.63 & 32.13 & 28.98 & FDGAN         & 23.15/0.921 & LPNet      & 33.61/0.958 & Stripformer & \textcolor{third}{33.08/0.962} & SMG           & 23.81/0.809 \\
ADFNet     & 34.77 & 32.22 & 29.06 & DehazeFormer  & \textcolor{second}{31.78/0.977} & DRSformer   & \textcolor{third}{38.14/0.983} & HI-Diff      & \textcolor{best}{33.33/0.964} & Retinexformer  & \textcolor{best}{25.16/0.845} \\
DGUNet  & 34.56 & 32.10 & 28.91 & Restormer     & 30.87/0.969 & Restormer  & 36.74/0.978 & MPRNet      & 32.66/0.959 & MIRNet        & \textcolor{second}{24.14/0.835} \\
Restormer  & 34.78 & 32.37 & 29.08 & NAFNet         & 30.98/0.970 & NAFNet      & 36.63/0.977 & Restormer   & 32.92/0.961 & Restormer     & 22.43/0.823 \\
FSNet      & 34.75 & 32.38 & 29.10 & FSNet          & 31.11/0.971 & FSNet       & 37.27/0.980 & FSNet        & \textcolor{second}{33.29/0.963} & DiffIR        & 23.15/0.828 \\
\midrule
AirNet         & \textcolor{third}{34.81} & \textcolor{third}{32.44} & 29.10 & AirNet        & 23.18/0.900 & AirNet       & 34.90/0.977 & AirNet       & 31.64/0.945 & AirNet        & 21.52/0.832 \\
IDR            & 34.78 & 32.42 & \textcolor{third}{29.13} & PromptIR      & 31.31/0.973 & PromptIR     & 37.04/0.979 & PromptIR     & 32.41/0.956 & PromptIR      & 22.97/0.834 \\
Perceive-IR    & \textcolor{second}{34.84} & \textcolor{second}{32.50} & \textcolor{second}{29.16} & Perceive-IR   & \textcolor{third}{31.65/0.977} & Perceive-IR  & \textcolor{best}{38.41/0.984} & Perceive-IR  & 32.83/0.960 & Perceive-IR   & 23.79/0.841 \\
\rowcolor{gray!10}
\textbf{Ours  }& \textcolor{best}{35.10} & \textcolor{best}{32.65} & \textcolor{best}{29.58} & \textbf{Ours}  & \textcolor{best}{32.80/0.979} & \textbf{Ours} & \textcolor{second}{38.15/0.984} & \textbf{Ours} & 31.45/0.946 & \textbf{Ours} & \textcolor{third}{24.03/0.827} \\
\bottomrule
\end{tabular}
\label{supp:tab:5t}
}
\end{table*}

\begin{figure}[!h]
    \centering
    \includegraphics[width=0.98\linewidth]{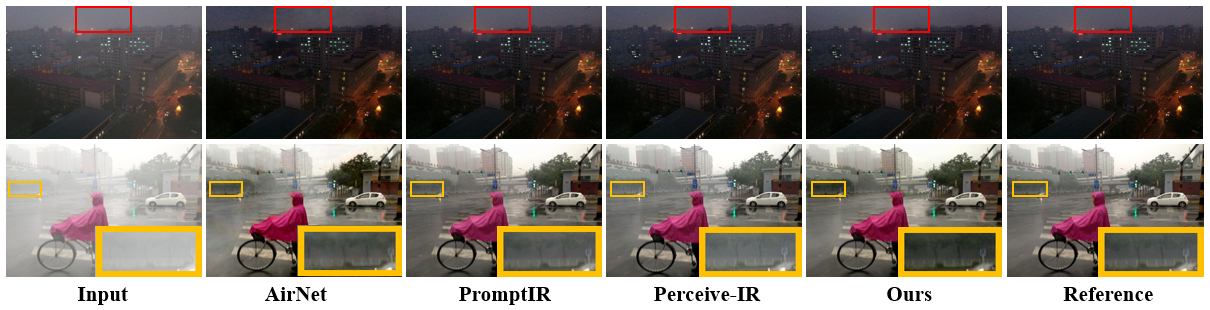}
    \caption{
    Visual Comparison of dehazed results between our IsoMamba and other all-in-one methods on the SOTS-outdoor.
    }
    \label{supp:fig:dehaze}
\end{figure}

\textbf{Visualization with all-in-one methods.} We provide qualitative comparisons on dehazing, deraining, and low-light enhancement using their corresponding test sets. Figure~\ref{supp:fig:dehaze} shows dehazing results under night-time (first row) and daytime (second row) conditions. In the night-time case, our method removes haze while introducing the least noise in the highlighted region (red box). In the daytime case, our approach better clears distant haze and restores finer structures, such as tree trunks and railings, with clearer textures in the zoomed-in areas.


\begin{figure}[!h]
    \centering
    \includegraphics[width=0.98\linewidth]{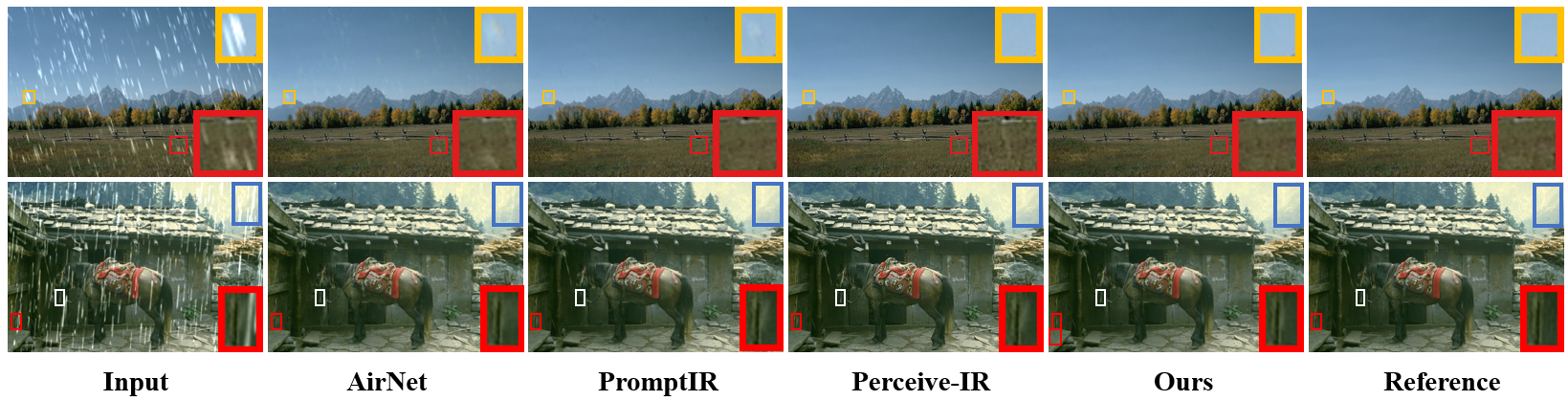}
    \caption{
    Visual Comparison of derained results between our IsoMamba and other all-in-one methods on the Rain100L.
    }
    \label{supp:fig:derain}
\end{figure}

As shown in Figure~\ref{supp:fig:derain}, our method demonstrates effective performance in both rain stripe removal and background restoration. In the first row, we successfully eliminated rain stripes within the red and yellow boxes. Simultaneously, in the second row, rain stripes were effectively removed from the blue box, whilst avoiding excessive generation of additional white artefacts when removing stripes from the red box. Meanwhile, within the white box, the background and rain stripes remain distinguishable, preserving the horse lead rope.

\begin{figure}[!h]
    \centering
    \includegraphics[width=0.98\linewidth]{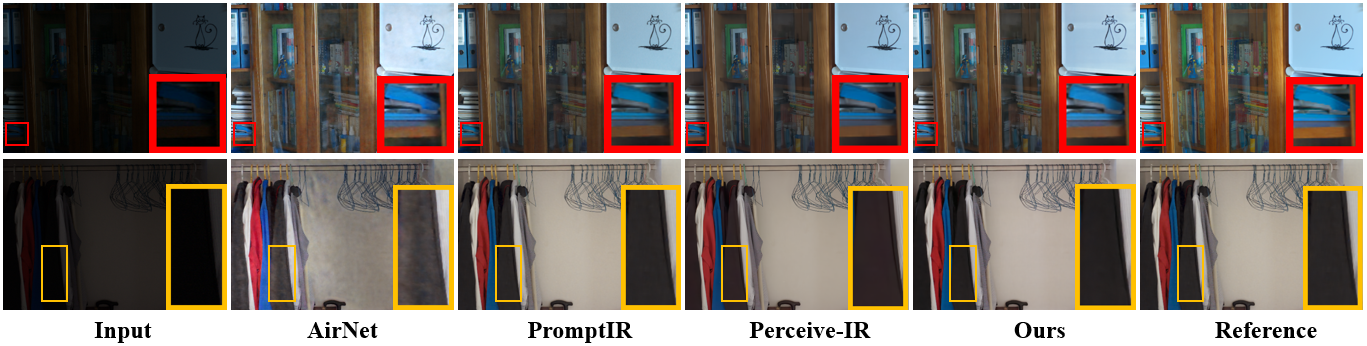}
    \caption{
    Visual Comparison of Low Light enhancement results between our IsoMamba and other all-in-one methods on the LoL.
    }
    \label{supp:fig:low_light}
\end{figure}

As illustrated in Figure~\ref{supp:fig:low_light}, our method restores brightness while maintaining reliable colour reproduction in low-light scenes. In the first row, it recovers the illumination in the red-bordered region and preserves the intended blue hue. In the second row, our approach restores dark tones without introducing artifacts, and it maintains a clean black appearance rather than drifting toward grey or brown colour shifts.

\section{Limitations and Future Work}
\label{supp:sec:limitation}
Our method is motivated by the mechanism-level diagnosis of State Contention, identifying that mechanism-agnostic serialization leads to conflicting update behaviors. We have substantiated this diagnosis through extensive hidden state visualizations and consistent quantitative improvements, utilizing the learned $\Delta$ as an effective, data-driven proxy for state isolation.
From a theoretical perspective, while our empirical evidence is robust, formally characterizing the bounds of error propagation caused by interleaved transitions remains an open frontier for the SSM community. We believe our findings pave the way for future work to develop a rigorous mathematical metric for "coupling intensity," linking the scan order directly to the stability analysis of recursive update equations.
In terms of application, while IsoMamba excels in handling color and haze, its performance on motion deblurring (e.g., GoPro) suggests room for improvement in modeling complex motion trajectories. Future iterations could explore integrating motion-aware priors into the clustering mechanism to further broaden the scope of restoration dynamics.

\bibliographystyle{plainnat}
\bibliography{aaai2027}
\end{document}